\documentclass[conference]{IEEEtran}
\IEEEoverridecommandlockouts
\usepackage{cite}
\usepackage{amsmath,amssymb,amsfonts}
\usepackage{algorithmic}
\usepackage{graphicx}
\usepackage{textcomp}
\usepackage{xcolor}
\usepackage{booktabs}
\usepackage{makecell}
\usepackage{multirow}
\usepackage{hyperref}
\def\BibTeX{{\rm B\kern-.05em{\sc i\kern-.025em b}\kern-.08em
    T\kern-.1667em\lower.7ex\hbox{E}\kern-.125emX}}
\begin{document}

\title{Deep Learning Foundation and Pattern Models: Challenges in Hydrological
Time Series}

\author{\IEEEauthorblockN{Junyang He}
\IEEEauthorblockA{\textit{Computer Science} \\
\textit{University of Virginia} \\
Charlottesville, VA, USA \\
fay5du@virginia.edu}
\and
\IEEEauthorblockN{Ying-Jung Chen}
\IEEEauthorblockA{\textit{College of Computing}\\
\textit{Georgia Institute of Technology} \\
Atlanta, GA, USA \\
yingjungcd@gmail.com}
\and
\IEEEauthorblockN{Alireza Jafari}
\IEEEauthorblockA{\textit{Computer Science} \\
\textit{University of Virginia} \\
Charlottesville, VA, USA \\
jrp5td@virginia.edu}
\and
\IEEEauthorblockN{Anushka Idamekorala}
\IEEEauthorblockA{\textit{Computer Science} \\
\textit{University of Virginia} \\
Charlottesville, VA, USA \\
anb5km@virginia.edu}
\and
\IEEEauthorblockN{Geoffrey Fox}
\IEEEauthorblockA{\textit{Biocomplexity Institute} \\
\textit{University of Virginia} \\
Charlottesville, VA, USA \\
vxj6mb@virginia.edu}}

\maketitle

\begin{abstract}
There has been active investigation into deep learning approaches for time series analysis, including foundation models. However, most studies do not address significant scientific applications. This paper aims to identify key features in time series by examining complex hydrology data. Our work advances computer science by emphasizing critical application features and contributes to hydrology and other scientific fields by identifying modeling approaches that effectively capture these features. Scientific time series data are inherently complex, involving observations from multiple locations, each with various time-dependent data streams and exogenous factors that may be static or time-varying and either application-dependent or purely mathematical. This research analyzes hydrology time series from the CAMELS and Caravan global datasets, which encompass rainfall and runoff data across catchments, featuring up to six observed streams and 209 static parameters across approximately 8,000 locations. Our investigation assesses the impact of exogenous data through eight different model configurations for key hydrology tasks. Results demonstrate that integrating exogenous information enhances data representation, reducing root mean squared error by up to 40\% in the largest dataset. Additionally, we present a detailed performance comparison of over 20 state-of-the-art pattern and foundation models. The analysis is fully open-source, facilitated by Jupyter Notebook on Google Colab for LSTM-based modeling, data preprocessing, and model comparisons. Preliminary findings using alternative deep learning architectures reveal that models incorporating comprehensive observed and exogenous data outperform more limited approaches, including foundation models. Notably, natural annual periodic exogenous time series contribute the most significant improvements, though static and other periodic factors are also valuable. This research serves as both an educational tool and benchmark resource.
\end{abstract}

\bibliographystyle{IEEEtran}

\section{Introduction}

\subsection{Spatio-temporal Series Datasets}

Scientific data is frequently represented as spatio-temporal series, where time series data are often influenced by geographical factors. The language of spatio-temporal series is used as a common application type, where the "series" can refer to any ordered sequential data points. These sequences can belong to any collection (bag), not restricted to Euclidean space-time, as long as sequences are labeled in some way and have properties that are consequent to the label. In the case of COVID-19 data \cite{cdcNationalCenter, cdcBehavioralRisk}, daily case / death statistics are grouped by location (e.g. city, county, country) and influenced by demographic characteristics of these locations \cite{covid-2021}. In the case of earthquake data \cite{usgsSearchEarthquake}, the earthquakes are grouped by 11 km $\times$ 11 km regions \cite{earthquake-2022}. Similarly, in the case of hydrology data, daily precipitation and streamflow are grouped by catchments and are affected by environmental attributes and locations of these sites. 

The spatio-temporal nature of scientific time series data means it is significantly influenced by the spatial variability of environmental factors. In recent years, computer science has seen the publication of hundreds of papers and more than fifty new deep learning models for time series analysis \cite{sciencefmhub,nixtla,ddz,survey2023,survey2024,Jafari2024Earthquake}, primarily focused on temporal dependencies. However, many of these models overlook the importance of incorporating application-specific environmental or exogenous data. This work addresses these gaps by emphasizing key considerations for developing time series models tailored to data-driven scientific applications.

\begin{figure}[htbp]
\includegraphics[width=\columnwidth]{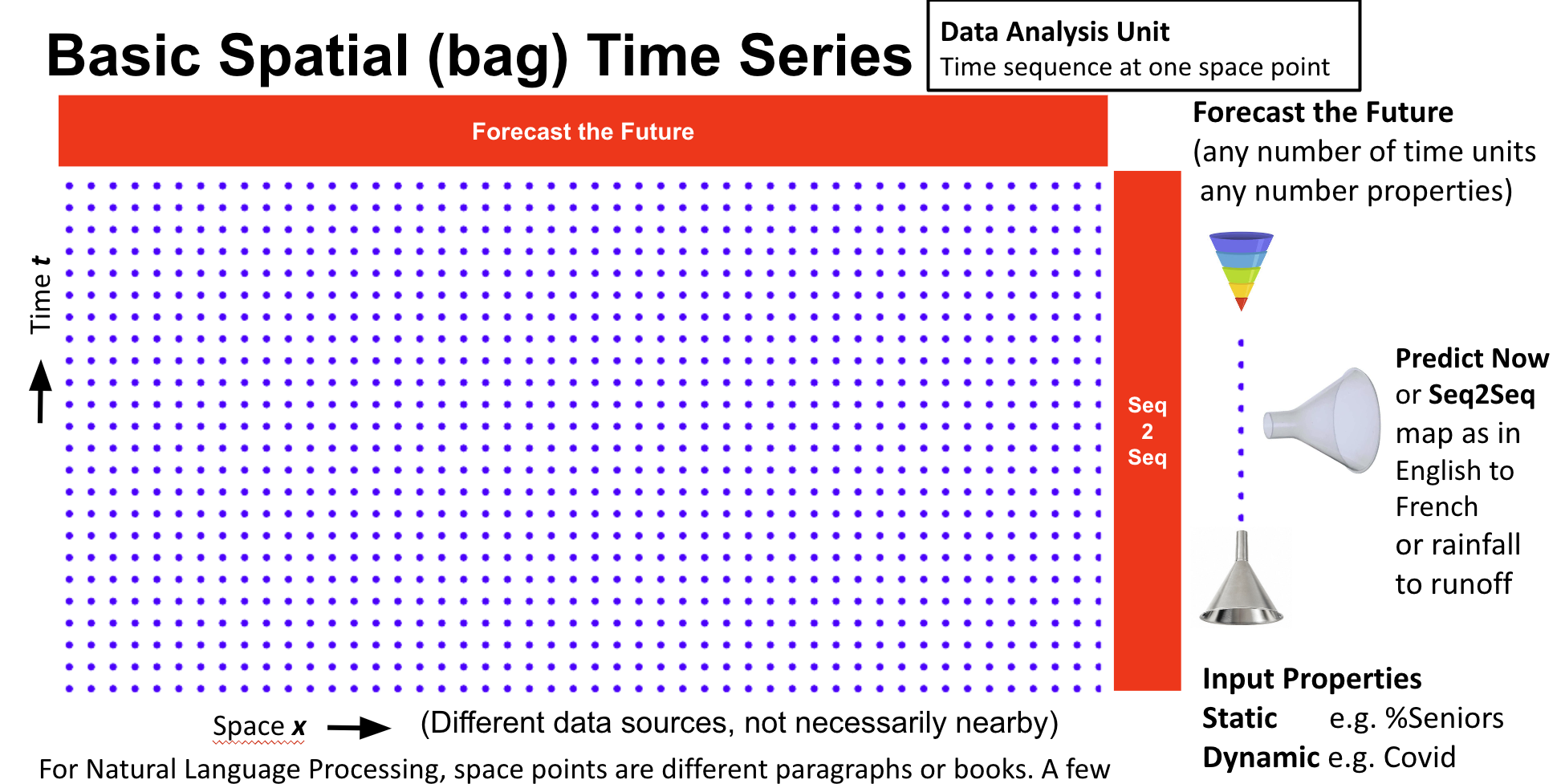}
\caption{Spatio-temporal series layout.}
\label{spatial_bag}
\end{figure}

Typically, the data in these contexts are recorded as space-time-stamped events. Fig~\ref{spatial_bag}. However, data can be converted into spatio-temporal series by binning in space and time. Comparing deep learning for time series with coupled ordinary differential equations for multi-particle systems motivates the use of an evolution operator to describe the time dependence of complex systems. Our research views deep learning applied to spatio-temporal series as a method for identifying the time evolution operator governing the behavior of complex systems.
Metaphorically, the training process uncovers hidden variables representing the system's underlying theory, similar to Newton's laws. Previous studies on COVID-19 \cite{covid-2021} and earthquakes \cite{earthquake-2022} show neural networks' ability to learn spatiotemporal dependencies in spatio-temporal series data. This work extends this approach to hydrology, demonstrating deep learning’s ability to model the rainfall-runoff process.

\subsection{Rainfall-Runoff Problem}
Rainfall-runoff modeling, a key challenge in hydrology, aims to model the physical process by which water on land surface (precipitation or snowmelt) moves to streams \cite{dingman2015physical}. As part of the hydrologic cycle, precipitation on land either evaporates, transpires, infiltrates to recharge groundwater, or becomes surface runoff entering a catchment \cite{dingman2015physical}. A catchment, or watershed, is an area where all precipitation collects and drains into a common outlet, such as a river, lake, or reservoir \cite{dingman2015physical}. Surface runoff contributes to streamflow, which is defined as the volumetric discharge that takes place in a stream or channel. Runoff $RO$ can be estimated with Equation. \ref{surface_runoff}, where $Q$ is streamflow, and $GW_{out}$ is ground-water outflow. Stream outflow $Q$ can be estimated with Equation. \ref{stream_outflow}, where $P$ is precipitation, $GW_{in}$ is ground-water inflow, $ET$ is evapotranspiration, $\Delta SM$ is the change in soil moisture. Eventually, this streamflow and ground water outflow reaches the ocean, where it evaporates, condenses into clouds, and returns as rainfall on land, completing the hydrologic cycle \cite{dingman2015physical}.

\begin{equation}
RO = Q + GW_{out}
\label{surface_runoff}
\vspace{-4mm}
\end{equation}

\begin{equation}
Q = P + GW_{in} - GW_{out} - ET - \Delta SM
\label{stream_outflow}
\vspace{2mm}
\end{equation}

Rainfall, like many natural phenomena, is periodic, with daily patterns that follow consistent seasonal cycles. While streamflow result from precipitation, it is also influenced by static environmental factors such as soil type, land cover, slope, etc. Based on this, we hypothesize that daily streamflow in a catchment can be predicted using a combination of daily meteorological forcing data, and the spatially and temporally distributed hydrologic, climatologic, geologic, pedologic, and land-use data \cite{dingman2015physical}. Additionally, a neural network can learn the seasonal patterns of these hydrological processes to produce accurate forecasts.

\subsection{Related Work}
Traditional rainfall-runoff modeling has typically focused on individual catchments. The first documented model, introduced in 1851, used linear regression to predict discharge from precipitation intensity and runoff \cite{mulvaney_1851}. Since then, scientific advancements have led to more sophisticated models based on mathematical formulas and physical laws. The advent of computers brought digital hydrological models. At a time when computers are expensive, slow, and low in memory, the Stanford Watershed Model \cite{Solr-29746} was proposed. It was seen as one of the first and most successful digital computer models \cite{beven_rainfall-runoff_2012}. As computers become more powerful, distributed models \cite{ABBOTT198645, moore_clark_pdm_1981} emerged, allowing for hydrological models to closely couple to geographical information systems for the input data \cite{beven_rainfall-runoff_2012}. Taking advantage of the number of parameters offered, these physical-based distributed models perform exceptionally well. However, the high computational cost to calibrate these parameters, and the limited availability of data hinder their use in large-scale forecasting applications \cite{overview_rainfall_runoff_2017}. 

Groundbreaking advancements in deep learning models and the publication of structured large-sample Hydrology dataset have overcome this limitation, enabling the study of nation or global scale rainfall-runoff modeling \cite{kratzert-lstm-hydrology-2018, shen_2018}. In the 1990s, Artificial Neural Network (ANN) based rainfall-runoff model was proposed \cite{hsu_artificial_neural_network_1995, tokar_a_sezin_rainfall-runoff_1999, ann-book-2013}. Although scientists were initially hesitant to embrace this novel “black box” approach due to the lack of extensive studies, deep learning based Hydrology models proved successful \cite{ann-book-2013}. Its exceptional capability in simulating complex non-linear systems is particularly advantageous for Hydrology modeling. In 2018, the focus of the field shifted towards Long Short-Term Memory (LSTM) based models, which excelled in learning sequential dependencies within time series data \cite{kratzert-lstm-hydrology-2018} \cite{hu-lstm-hydrology-2018}. These models have shown great success in large-scale hydrological time series predictions. Further studies explored the interpretability of such LSTM models within physics context \cite{Kratzert2019}. Since then, an open source library for the LSTM based rainfall-runoff model was published \cite{kratzert_neuralhydrology_2022}.

\section{Data And Methods}

\subsection{Datasets selection}

Hydrology data comprises of both time series and static exogenous features. It falls within the scope of spatio-temporal series since all time series properties (eg. mean temperature, streamflow) are collected and organized by catchment. Hydrology data is collected by gauges, which are stations that collect measurements at each catchment. Static attributes for each catchment refers to environmental conditions (eg. dominant land cover, soil aridity), as well as spatial extent and locations (eg. coordinates).

Recent deep learning studies on Hydrology have been driven by the advent of CAMELS (Catchment Attributes and Meteorology for Large-sample Studies) datasets, which established a standard for organizing big Hydrology data at different nations across the globe. The first CAMELS dataset \cite{camels-us-2017}, initially proposed with 671 catchments in the U.S., benchmarked the types of static and time series properties necessary for large sample hydrology datasets containing hundreds of catchments or more. The high dimensionality of static data along with the 20-year duration of daily time series data made it suitable for nation-scale Hydrology modeling using neural network models. Since then, CAMELS-standard datasets have been published for countries including the United Kingdom \cite{camels-gb-2020}, Chile \cite{camels-cl-2018}, Australia \cite{camels-aus-2021}, Brazil \cite{camels-br-2020}, Switzerland \cite{camels-ch-2023}, Sweden \cite{camels-se-2024}, France \cite{camels-fr-2022}, etc.

\subsection{Three-Nation Combined CAMELS Data}
Similarities shared by CAMELS-standard national datasets allow for the combination of national datasets into a large global dataset, which can be used for global-scale training. In this study, the experimental dataset was produced by combining CAMELS data from three nations: the United States \cite{camels-us-2017}, United Kingdom \cite{camels-gb-2020}, and Chile \cite{camels-cl-2018}. These datasets were selected as they provide the earliest available CAMELS-structured data during the data preprocessing phase of this study. We select static features and time series targets that are shared across the three datasets prior to combination.

\renewcommand{\arraystretch}{1.5}
\begin{table*}[htbp]
\caption{Three-Nation Combined CAMELS Data Source}
\begin{center}
\begin{tabular}{ c c c }
\toprule
&\textbf{Forcing $^{\mathrm{a}}$} &\textbf{Streamflow} \\
\hline
\textbf{CAMELS-US} & Maurer \cite{maurer-2013} & USGS \cite{usgs-2012}\\
\hline
\textbf{CAMELS-GB} & CEH-GEAR \cite{ceh-gear-2015} & NRFA \cite{nrfa-2004} \\
& CHESS-met \cite{chess-met-2017} & \\
\hline
\textbf{CAMELS-CL} &  CR2MET \cite{crmet-2023} & CR2 \cite{cr-streamflow} \\
\bottomrule
\multicolumn{3}{l}{$^{\mathrm{a}}$ Time series targets such as precipitation and temperature.}
\end{tabular}
\label{camels_data_source}
\end{center}
\end{table*}

Although CAMELS-US, CAMELS-GB, CAMELS-CL follow the same standards, the exact choice of static attributes they include vary slightly, prohibiting simple concatenation of data. For instance, the “Land Cover” section of CAMELS-US only contains “\% cover of forest” data for each catchment, while that section of CAMELS-GB contains the percent cover of all plantation types like woodland, crops, shrublands, etc. Without further details on the data measurement process for these properties, arithmetic manipulations to forge “percent cover of forest” for the CAMELS-GB dataset from the given “percent cover of woodland, crops, shrublands, etc” cannot be performed. To address this issue, the three-nation combined CAMELS dataset used in this study only contains static properties shared across the US, GB, and CL datasets. The processed dataset includes 1858 catchments, 3 dynamic, and 29 static variables.

\subsection{CAMELS Preprocessing}
For dynamic variables, we selected the common interval of 7,031 days, spanning from October 2, 1989, to December 31, 2008. The start date of a water year, as defined by the U.S. Geological Survey, is October 1st \cite{dingman2015physical}. To account for time differences between the nations, we shifted the training data by one day, beginning on October 2nd. An analysis of NaNs (missing values) in data revealed no NaN values in time series data, yet some exists in the static features. Since missing data constitutes only a small percentage of total static data and occurs miscellaneously, it is filled with the mean value of that attribute from all catchments. Categorical variables are only present in the form of months from January to December. We encode them with ordinal encoding between 0 and 1 to preserve the natural order of months.

\subsection{Caravan}
Published in 2023, the Caravan dataset \cite{caravan-2023}, consists of seven preprocessed CAMELS-standard national datasets that contain identical static exogenous features and time series properties. Caravan aggregates data from 6,830 catchments across 16 nations spanning four continents, making it ideal for global rainfall-runoff modeling with large-scale hydrological data.

\begin{table*}[htbp]
\caption{Caravan Sub-datasets}
\begin{center}
\begin{tabular}{ c c }
\hline
\textbf{Sub-dataset} &\textbf{Catchments}  \\
\hline
\textbf{CAMELS (US) \cite{camels-us-2017}} & 482\\
\textbf{CAMELS-AUS \cite{camels-aus-2021}} & 150\\
\textbf{CAMELS-BR \cite{camels-br-2020}} & 376\\
\textbf{CAMELS-CL \cite{camels-cl-2018}} & 314\\
\textbf{CAMELS-GB \cite{camels-gb-2020}} & 408\\
\textbf{HYSETS (North America) \cite{hysets_2020}} & 4621\\
\textbf{LamaH-CE \cite{lamah-2021}} & 479\\
\hline
\end{tabular}
\label{caravan}
\end{center}
\end{table*}

The Caravan dataset is not constructed by identifying common static and time series properties shared by previously published versions of each sub-dataset. Instead, it is derived from data sources that differ significantly from those used in the original CAMELS datasets for various nations. While the catchments included in the Caravan sub-datasets are the same as those in the original versions, the actual data is sourced from global hydrological sources rather than the local sources used in the original publications. This shift to global sources enables the collection of more standardized and abundant data, making it suitable for global-scale comparative hydrological studies—one of the key motivations behind the creation of Caravan. However, it is important to note that data from global sources may vary slightly from local sources, which are often more precise. For example, in the CAMELS-US sub-dataset within Caravan, all forcing data is sourced from ERA5-Land \cite{era5-2021}, streamflow data from GSIM \cite{gsim-p1-2018} \cite{gsim-p2-2018}, and static properties from HydroATLAS \cite{hydroatlas-2019}. In contrast, the original CAMELS-US \cite{camels-us-2017} dataset included static and time series data from three sources: NLDAS \cite{nldas-2012}, Maurer \cite{maurer-2013}, and DayMet \cite{daymet-2014}, with streamflow data from USGS \cite{usgs-2012}. After comparison, only the streamflow data remains consistent between CAMELS-US in the original publication and in Caravan. The correlation coefficient for precipitation time series data between different data sources is presented in \ref{camels_caravan_data_comparison}. Nevertheless, the Caravan paper \cite{caravan-2023} addresses this concern, noting that the correlations between Caravan and each of the three CAMELS-US data products are not consistently lower than the correlations within the individual CAMELS-US data products.

\renewcommand{\arraystretch}{1.5}
\begin{table}[htbp]
\vspace{-2mm}
\caption{CAMELS-US Caravan Comparison}
\begin{center}
\begin{tabular}{ c|c|c|c }
&\textbf{Maurer} &\textbf{Daymet} &\textbf{NLDAS} \\
\hline
\textbf{Caravan}& 0.71532 & 0.60720 & 0.75441\\
\end{tabular}
\label{camels_caravan_data_comparison}
\end{center}
\vspace{-2mm}
\end{table}

\section{Methods}

\subsection{Long Short-Term Memory}
The Long Short-Term Memory (LSTM) network \cite{lstm-1997} is a specialized variant of a Recurrent Neural Network (RNN) designed to address the vanishing gradient problem through its unique memory cell structure. In an LSTM block (as shown in the figure below), the cell state (denoted by $C$) serves as long-term memory. Minimal weight updates to the cell state during backpropagation effectively mitigate the vanishing gradient problem, making LSTM particularly well-suited for time series tasks involving long sequences of data. Additionally, the hidden state (denoted by $h$) serves as short-term memory. Weights (denoted by $W$) and biases (denoted by $b$) are applied to the input and passed through sigmoid and tanh activation functions. The output from the hidden state is then used to update the cell state, while the updated cell state, in turn, informs the hidden state during the final stage of computation within the LSTM block.

\begin{figure}[htbp]
\includegraphics[width=\columnwidth]{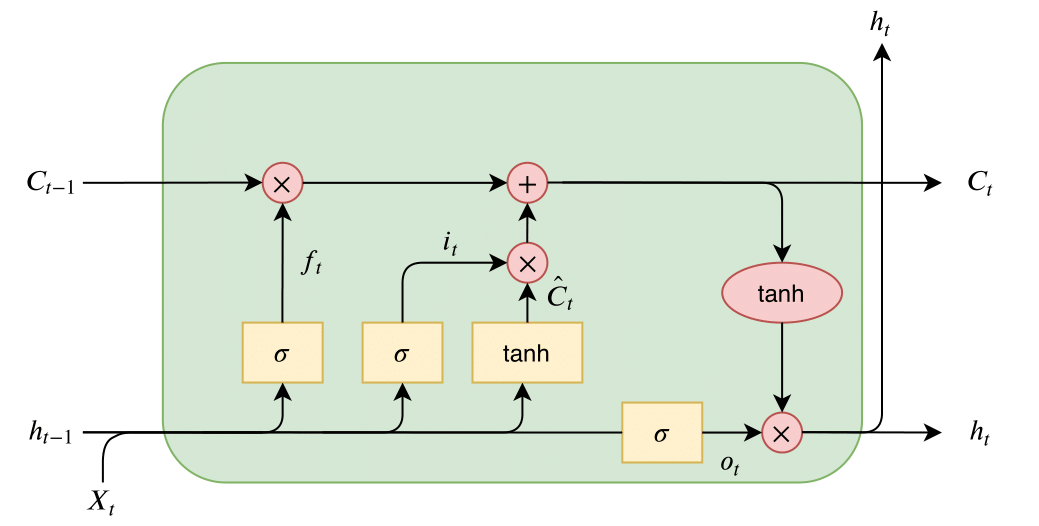}
\caption{LSTM network.}
\label{lstm}
\end{figure}

\textbf{Forget gate.} Initially, the input $X_t$ is combined with the previous hidden state $h_{t-1}$ through the forget gate, which determines how much of the long-term memory is preserved. As shown in Equation \ref{forget_gate}, $f_t$ represents the forget gate value, $W_f$ denotes the input weights for the forget gate, $U_f$ refers to the recurrent weights for the forget gate, and $b_f$ is the bias term for the forget gate.

\textbf{Input gate.} Next, the input $X_t$ passes through the input gate, which determines the specific information to be incorporated into the long-term memory. As shown in Equation \ref{input_gate}, $i_t$ represents the input gate value, $\hat{C}_t$ denotes the temporary cell state value, $W_i$ represents the input weights for the input gate, $U_i$ refers to the recurrent weights for the input gate, and $b_i$ is the bias term for the input gate.

\textbf{Output gate.} Lastly, the input $X_t$ passes through the output gate, which updates the short-term memory. As shown in Equation \ref{output_gate}, $O_t$ represents the output gate value, $W_o$ is the input weights for the output gate, $U_o$ is the recurrent weights for the output gate, and $b_o$ is the bias term for the output gate.

\textbf{Cell state and hidden state.} The outputs from the forget gate, input gate, and output gate are applied to the previous cell state and hidden state to calculate the new "long-term" and "short-term" memory values. As shown in Equations \ref{cell_state} and \ref{hidden_state}, $C_t$ represents the updated cell state (long-term memory), and $h_t$ represents the updated hidden state (short-term memory).

\begin{equation}
f_t = sigmoid(W_f \times x_t + U_f \times h_{t-1} + b_f)
\label{forget_gate}
\vspace{-2mm}
\end{equation}

\begin{equation}
i_t = sigmoid(W_i \times x_t + U_i \times h_{t-1} + b_i) 
\label{input_gate}
\vspace{-2mm}
\end{equation}

\begin{equation}
\hat{C}_t = tanh(W_i \times x_t + U_i \times h_{t-1} + b_i)
\label{temporary_cell_state}
\vspace{-2mm}
\end{equation}

\begin{equation}
O_t = sigmoid(W_o \times x_t + U_o \times h_{t-1} + b_o)
\label{output_gate}
\vspace{-2mm}
\end{equation}

\begin{equation}
C_t = C_{t-1} \times f_t + i_t \times \hat{C}_t
\label{cell_state}
\vspace{-2mm}
\end{equation}

\begin{equation}
h_t = O_t \times tanh(C_t)
\label{hidden_state}
\end{equation}

\subsection{Model Setup}
\begin{figure}[htbp]
\includegraphics[width=\columnwidth]{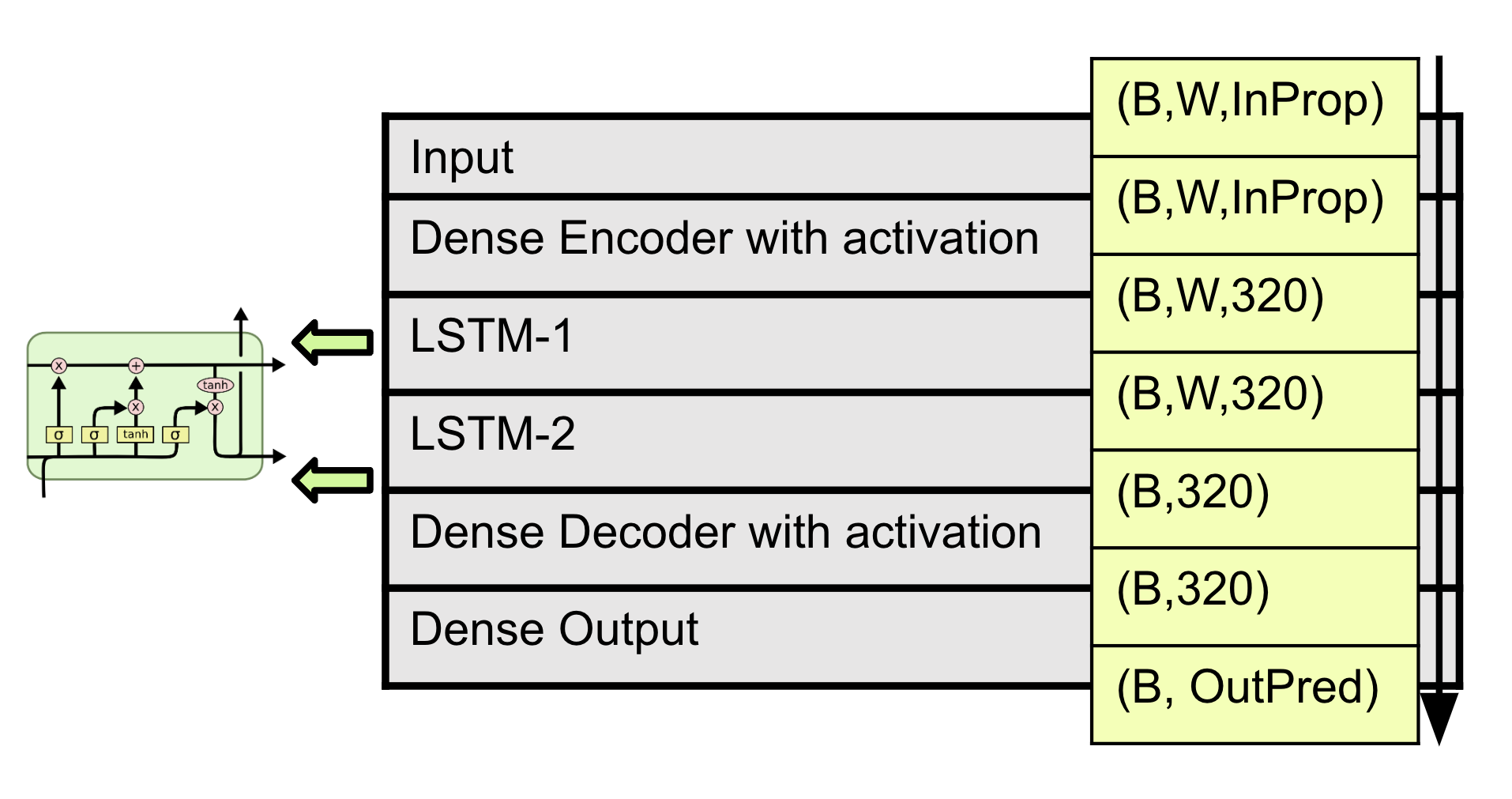}
\caption{CAMELS-US model architecture and layer size.}
\label{model_architecture}
\end{figure}

The LSTM model used in this study is implemented using the Tensorflow framework, which allows for customization of layer parameters. The baseline model architecture is shown in Fig.~\ref{model_architecture}. Detailed activation function setup is shown in Tab.~\ref{activation}. To prevent overfitting, a dropout rate of 20\% is applied to the layers. 

\renewcommand{\arraystretch}{1}
\begin{table}[htbp]
\caption{Model Activation Setup}
\begin{center}
\begin{tabular}{ c c }
\toprule
\textbf{Dense Encoder Activation} & SELU\\
\textbf{LSTM Recurrent Activation} & Sigmoid\\
\textbf{LSTM Layer Activation} & SELU\\
\textbf{Dense Decoder Activation} & SELU\\
\bottomrule
\end{tabular}
\label{activation}
\end{center}
\vspace{-6mm}
\end{table}

\subsection{Model Training and Evaluation}
Inputs to the model can be classified into known inputs and observed inputs \cite{earthquake-2022}. Known inputs refer to features that are known in both the past and future. In this hydrology study, static known inputs are the exogenous features such as climatic signatures, hydrologic signatures, and catchment topography. Additionally, time series known inputs exist in the form of seasonal patterns hidden in all time series features. Observed inputs are features known only for the past time periods but unknown in the future. In hydrology, those features include precipitation, temperature, and streamflow. Model output, known as targets, are the time-dependent predicted properties. In this study, predicted targets are precipitation, temperature, and streamflow.

The input data is divided into batches with a sequence length of 21 days, selected after testing various other lengths, including 7, 14, and 365 days. The 3-week sequence length was chosen to effectively capture subtle hydrological patterns. The total number of batches processed during one epoch is calculated using Equation \ref{batch_per_epoch}, while the batch size is determined by Equation \ref{batch_size}. In these equations, $Day_{total}$ represents the total duration of the input data in days, $l_{seq}$ denotes the sequence length, $\# Gauges$ refers to the total number of gauges in the input data, and $\# Input\ properties$ indicates the total number of input features, including both static and time series properties.

\begin{equation}
\# batch/epoch = Day_{total} - l_{seq} + 1
\label{batch_per_epoch}
\vspace{-6mm}
\end{equation}

\begin{equation}
S_{batch} = l_{seq} \times \# Gauges \times \#Input\ properties
\label{batch_size}
\end{equation}

\textbf{Symbolic Window.} A key challenge encountered during LSTM model training was the space complexity associated with handling the training data. Under the previously mentioned batch size configuration, storing all batches prior to training requires RAM space of $O(S_{batch} \times \#batch/epoch)$. To address this issue, we leverage a symbolic window to dynamically generate batches during each epoch of training. Instead of storing the entire training set, including all batches, before training, we only store the original input data with a space complexity of $O(Day_{total} \times \#Gauges \times \#Input\ properties)$. During each epoch, we track the start index $i$ for the current batch and extract the corresponding data for the batch from the input data using $Data[day_{i} : day_{i+l_{seq}}]$. This batch is then trained, and the index $i$ is incremented before proceeding to extract the next batch. This process is repeated $\#batch/epoch$ times to complete a full epoch of training. Note: This technique is environment-dependent and may not be applicable to all frameworks.

\textbf{Spatial and Temporal Encodings.} Like many fields of science, hydrology time series data exhibit strong seasonal patterns. It can reasonably be assumed that, at a specific gauge location, the precipitation and streamflow in October of one year will be similar to those recorded in October of the previous year. Furthermore, research has identified the approximate water residence times in various reservoirs, such as rivers, lakes, soil, and the atmosphere \cite{pidwirny_hydrologic_cycle}. To effectively capture these known dependencies within the time series properties, spatial and temporal encodings are incorporated into the model during training.
\begin{enumerate}
\item \textbf{Linear Space:} a linear function with length equaling total number of catchments in input data.

\item \textbf{Linear Time:} a linear function with length equaling total number of days in input time series.

\item \textbf{Annual Fourier Time:} a basic sine and cosine function with period equaling one year.

\item \textbf{Extra Fourier Time:} basic sine and cosine functions with period equaling 8, 16, 32, 64, 128 days.

\item \textbf{Legendre Time:} Legendre functions of degree 2, 3, and 4 with range equaling total number of days in input time series.
\end{enumerate}

\textbf{Spatial Validation.} We employ location-based rather than temporal-based validation as the catchments in the CAMELS and Caravan datasets are uncorrelated. The training and validation datasets are randomly selected on an 8:2 ratio by location. For example, out of the 671 catchments included in CAMELS-US dataset, 537 catchments are used for training and 134 catchments are used for validation. This decision is made given the extensive number of catchments available in the datasets used in this study. 

\textbf{Training Details.} Each benchmark model is trained for 120 successful epochs, a value empirically selected to balance predictive accuracy and computational efficiency. An epoch is considered successful if either the training or validation loss improves relative to the previous epoch. We employ the Adam optimizer \cite{kingma2017adammethodstochasticoptimization} with a learning rate of 0.001. Final model evaluation is performed using the checkpoint from the 120th epoch if the loss curve indicates no signs of overfitting. All training is performed on an NVIDIA A100 GPU via Google Colab. On average, training a single epoch with the CAMELS-US dataset \cite{camels-us-2017} requires approximately 30 minutes, while an epoch with the Caravan global dataset takes about one hour. With the same setup, GPU memory usage peaks at 1811MB for CAMELS-US run and 9431MB for Caravan global run.

\textbf{Evaluation Metrics.} Model fit is quantified using root mean squared error (RMSE) and normalized Nash-Sutcliffe Efficiency (NNSE) scores \cite{nnse-2012}. The NNSE score is calculated using Equation \ref{nnse}, where $T$ represents the total number of days, $Q_m^t$ represents the modeled discharge on day $t$, $Q_o^t$ represents the observed discharge on day $t$, and $\overline{Q}_o$ is the mean observed discharge over $T$ days. NNSE values range from 0 (poor performance) to 1 (perfect fit), with a score of 0.5 indicating that the model's predictions are equivalent to the time-averaged mean of the observations. In this study, the NNSE value is calculated for each gauge over time $T$, and the average NNSE across all gauges in the input data is reported.
\begin{equation}
NNSE = \frac{\Sigma_{t=1}^T (Q_m^t - \overline{Q}_o)^2}{\Sigma_{t=1}^t (Q_o^t - Q_m^t)^2 + (Q_o^t - \overline{Q}_o)^2}
\label{nnse}
\end{equation}

\section{LSTM Benchmark Runs}
\subsection{Experiment Setup}
All experiments are conducted using the model architecture shown in Fig. \ref{model_architecture} and the activation functions listed in Table \ref{activation}. The benchmark runs utilize \textbf{Linear Space}, \textbf{Linear Time}, and \textbf{Annual Fourier Time} encodings. We leverage multivariate time series forecasting \cite{doi:10.1080/02626669509491401}. Streamflow time series data is not trained but predicted as target.

\subsection{Three-Nation Combined Benchmark Run}
This run evaluates model performance on the combined CAMELS dataset from three nations: the US, UK, and Chile. The input data include static properties that are common across all three regions, whereas predicted targets consist of precipitation, mean temperature, and streamflow. Results are demonstrated in Table \ref{three-nation_fit}.

\renewcommand{\arraystretch}{1.4}
\begin{table}[htbp]
\caption{CAMELS Three-Nations Combined Benchmark Run}
\begin{center}
\begin{tabular}{ c c c c }
\toprule
& &\textbf{RMSE} &\textbf{NNSE} \\
\hline
                                                                        & Train & 0.002764  & 0.847 \\ \cline{2-4}
\multirow{-2}{*}{\textbf{Precipitation}}                                & Val   & 0.003200  & 0.836 \\ \midrule
                                                                        & Train & 0.000212  & 0.933 \\ \cline{2-4}                                                   
\multirow{-2}{*}{\makecell[l]{\textbf{Mean}\\ \textbf{Temperature}}}    & Val   & 0.000356  & 0.933 \\ \midrule
                                                                        & Train & 0.000432  & 0.697 \\ \cline{2-4}
\multirow{-2}{*}{\textbf{Streamflow}}                                   & Val   & 0.000613  & 0.654 \\ \midrule
                                                                        & Train & 0.003438  & - \\ \cline{2-4}
\multirow{-2}{*}{\textbf{Total}}                                        & Val   & 0.004325  & - \\
\bottomrule

\end{tabular}
\label{three-nation_fit}
\end{center}
\vspace{-2mm}
\end{table}

\subsection{CAMELS Caravan US Benchmark Runs}
The runs compare model performance between the original CAMELS-US dataset and the US sub-dataset within the Caravan dataset. The CAMELS-US model is trained using selected static and time series data from the original CAMELS-US dataset, while the Caravan-US model is trained using selected static and time series data from the US sub-dataset within Caravan. For both runs, we utilize the same time series features—precipitation and mean temperature—for training and predict the same targets: precipitation, mean temperature, and streamflow. Results are presented in Table \ref{camels_caravan_comparison}.

\setlength{\tabcolsep}{4.5pt}
\renewcommand{\arraystretch}{1.4}
\begin{table}[htbp]
\caption{CAMELS Caravan US Comparison}
\begin{center}
\small
\begin{tabular}{l l cc cc}
\toprule
                                            &           & \multicolumn{2}{c}{\textbf{CAMELS-US}}    & \multicolumn{2}{c}{\textbf{Caravan US}} \\ \cmidrule(rl){3-6}
                                            &           & RMSE       & NNSE  & RMSE       & NNSE  \\ 
                                            \cmidrule(rl){3-3} \cmidrule(rl){4-4} \cmidrule(rl){5-5} \cmidrule(rl){6-6}  
                                            & Train     & 0.0592  & 0.820 & 0.0540  & 0.851 \\ \cline{2-6} 
\multirow{-2}{*}{\textbf{Precipitation}}    & Val       & 0.0599  & 0.819 & 0.0656  & 0.801 \\ \midrule
                                            & Train     & 0.0166  & 0.961 & 0.0216  & 0.967 \\ \cline{2-6} 
\multirow{-2}{*}{\makecell[l]{\textbf{Mean}\\ \textbf{Temperature}}}   & Val       & 0.0168 & 0.960 & 0.0239  & 0.963 \\ \midrule
                                            & Train     & 0.0169  & 0.806 & 0.0229  & 0.814 \\ \cline{2-6} 
\multirow{-2}{*}{\textbf{Streamflow}}       & Val       & 0.0172  & 0.812 & 0.0309  & 0.703 \\ \midrule
                                            & Train     & 0.0641  & -     & 0.0631  & - \\ \cline{2-6} 
\multirow{-2}{*}{\textbf{Total}}            & Val       & 0.0648  & -     & 0.0774  & - \\ 
\bottomrule

\end{tabular}
\label{camels_caravan_comparison}
\end{center}
\vspace{-4mm}
\end{table}

\subsection{Caravan PCA Runs}
This experiment explores the use of Principal Component Analysis (PCA) \cite{WOLD198737} to reduce the dimensionality of static properties in the Caravan input data. The Caravan sub-datasets contain over 200 static properties, nearly seven times more than the number of static input properties used in the original CAMELS studies. While this extensive range of static properties enhances model training, it significantly increases computational demands and GPU usage. To address this, we apply PCA, a widely adopted dimensionality reduction technique, to reduce the number of static input features to a level comparable with that in the CAMELS studies. In this experiment, we set the explained variance threshold to 90\%, resulting in the reduction of static input features to approximately 30.

The first part of the experiment assesses the effect of PCA on model trained on US sub-dataset within the Caravan dataset, representing a small-scale input. The second part of the experiment examines the effect of PCA on model trained on North America regional data (HYSETS) \cite{hysets_2020} within the Caravan dataset, representing a large-scale input.

Results, shown in Table \ref{caravan_us_pca_test} and Table \ref{caravan_hysets_pca_test}, indicate that the models trained with static properties obtained from PCA perform comparably to the models trained with original static properties, demonstrating that the reduction in input static dimensionality does not significantly compromise model accuracy. These findings validate PCA as an effective approach for train hydrology time series models with high static dimensionality.

\renewcommand{\arraystretch}{1.4}
\begin{table}[htbp]
\caption{Caravan US PCA Experiment}
\begin{center}
\begin{tabular}{l l cc cc}
\toprule
                                            &           & \multicolumn{2}{c}{\textbf{Original Static}}    & \multicolumn{2}{c}{\textbf{PCA Static}} \\ \cmidrule(rl){3-6}
                                            &           & MSE       & NNSE  & MSE       & NNSE  \\ 
                                            \cmidrule(rl){3-3} \cmidrule(rl){4-4} \cmidrule(rl){5-5} \cmidrule(rl){6-6}  
                                            & Train     & 0.002920  & 0.851 & 0.002994  & 0.848 \\ \cline{2-6} 
\multirow{-2}{*}{\textbf{Precipitation}}    & Val       & 0.004307  & 0.801 & 0.004264  & 0.800 \\ \midrule
                                            & Train     & 0.000468  & 0.967 & 0.000468  & 0.967 \\ \cline{2-6} 
\multirow{-2}{*}{\makecell[l]{\textbf{Mean}\\ \textbf{Temperature}}}   & Val       & 0.000573 & 0.963 & 0.000548  & 0.965 \\ \midrule
                                            & Train     & 0.000525  & 0.814 & 0.000569  & 0.799 \\ \cline{2-6} 
\multirow{-2}{*}{\textbf{Streamflow}}       & Val       & 0.000955  & 0.703 & 0.000989  & 0.703 \\ \midrule
                                            & Train     & 0.003982  & -     & 0.004075  & -     \\ \cline{2-6} 
\multirow{-2}{*}{\textbf{Total}}            & Val       & 0.005987  & -     & 0.005871  & -     \\ \bottomrule
\end{tabular}
\label{caravan_us_pca_test}
\end{center}
\vspace{-2mm}
\end{table}

\setlength{\tabcolsep}{4.5pt}
\renewcommand{\arraystretch}{1.4}
\begin{table}[htbp]
\caption{Caravan Hysets PCA Experiment}
\begin{center}
\small
\begin{tabular}{l l cc cc}
\toprule
                                            &           & \multicolumn{2}{c}{\textbf{Original Static}}    & \multicolumn{2}{c}{\textbf{PCA Static}} \\ \cmidrule(rl){3-6}
                                            &           & RMSE       & NNSE  & RMSE       & NNSE  \\ 
                                            \cmidrule(rl){3-3} \cmidrule(rl){4-4} \cmidrule(rl){5-5} \cmidrule(rl){6-6}  
                                            & Train     & 0.0521  & 0.835 & 0.0530  & 0.830 \\ \cline{2-6} 
\multirow{-2}{*}{\textbf{Precipitation}}    & Val       & 0.0544  & 0.826 & 0.0553  & 0.821 \\ \midrule
                                            & Train     & 0.0193  & 0.968 & 0.0196  & 0.967 \\ \cline{2-6} 
\multirow{-2}{*}{\makecell[l]{\textbf{Mean}\\ \textbf{Temperature}}}   & Val       & 0.0196 & 0.966 & 0.0199  & 0.965 \\ \midrule
                                            & Train     & 0.0248  & 0.825 & 0.0264  & 0.813 \\ \cline{2-6} 
\multirow{-2}{*}{\textbf{Streamflow}}       & Val       & 0.0296  & 0.798 & 0.0199  & 0.799 \\ \midrule
                                            & Train     & 0.0589  & -     & 0.0601  & -     \\ \cline{2-6} 
\multirow{-2}{*}{\textbf{Total}}            & Val       & 0.0624  & -     & 0.0627  & -     \\ \bottomrule
\end{tabular}
\label{caravan_hysets_pca_test}
\end{center}
\vspace{-4mm}
\end{table}

\subsection{Caravan Global Benchmark Run}
 This study provides insights into the model's applicability to large-scale global hydrology datasets. The input data is compiled by concatenating all seven Caravan sub-datasets, which encompass catchments from four continents. As shown in Table \ref{caravan_global_fit}, accuracy decreases slightly compared to small-scale, individual nation fits (Table \ref{camels_caravan_comparison}), yet overall model performance remains relatively high, demonstrating its robustness on global datasets.
\renewcommand{\arraystretch}{1.4}
\begin{table}[htbp]
\caption{Caravan Global Fit}
\begin{center}
\begin{tabular}{ c c c c }
\toprule
& &\textbf{RMSE} &\textbf{NNSE} \\
\midrule
                                                                        & Train & 0.0558  & 0.809 \\ \cline{2-4}
\multirow{-2}{*}{\textbf{Precipitation}}                                & Val   & 0.0568  & 0.808 \\ \midrule
                                                                        & Train & 0.0189  & 0.953 \\ \cline{2-4}                                                   
\multirow{-2}{*}{\textbf{Mean Temperature}}    & Val   & 0.0189  & 0.952 \\ \midrule
                                                                        & Train & 0.0260  & 0.781 \\ \cline{2-4}
\multirow{-2}{*}{\textbf{Streamflow}}                                   & Val   & 0.0272  & 0.768 \\ \midrule
                                                                        & Train & 0.0630  & -     \\ \cline{2-4}
\multirow{-2}{*}{\textbf{Total}}                                        & Val   & 0.0643  & -     \\
\bottomrule
\multicolumn{4}{l}{*Model trained with PCA static features.} \\

\end{tabular}
\label{caravan_global_fit}
\end{center}
\vspace{-6mm}
\end{table}

\section{Static Properties and Spatial Temporal Encodings Experiment}

\begin{figure}[htbp]
\includegraphics[width=\linewidth]{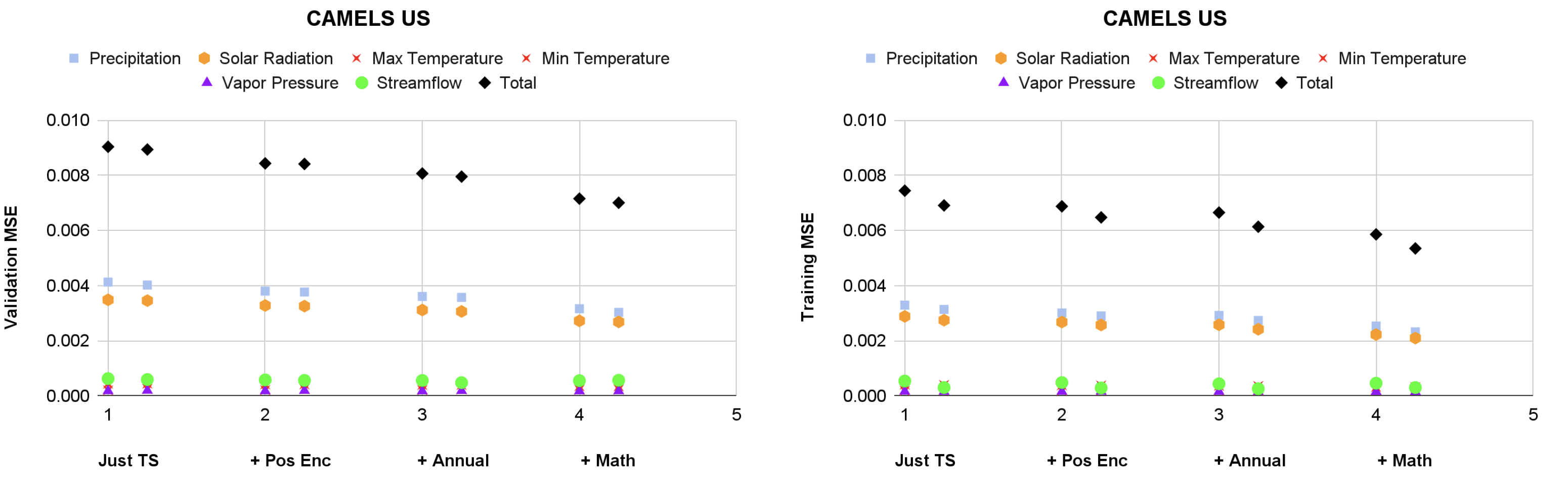}
\vspace{-4mm}
\caption{CAMELS US experiment (RMSE loss). Each run configuration is utilized in 2 runs (left: static features not trained, right: static features trained).}
\label{camels_us_static_encodings}
\vspace{3mm}

\includegraphics[width=\linewidth]{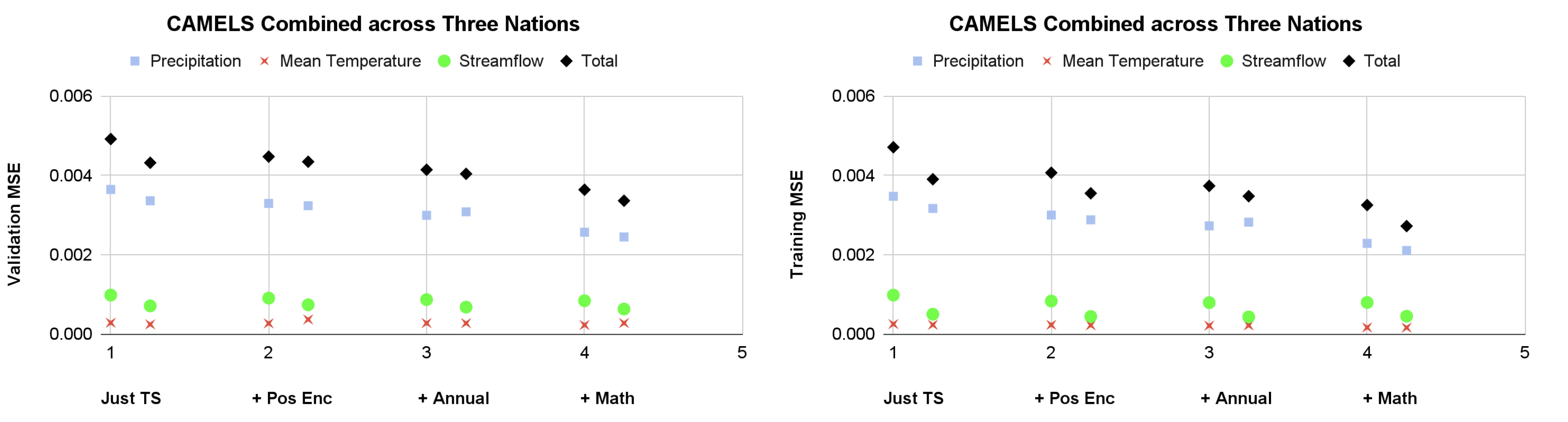}
\vspace{-4mm}
\caption{CAMELS three-nations combined experiment (RMSE loss). Each run configuration is utilized in 2 runs (left: static features not trained, right: static features trained).}
\label{camels_combined_static_encodings}
\vspace{3mm}

\includegraphics[width=\linewidth]{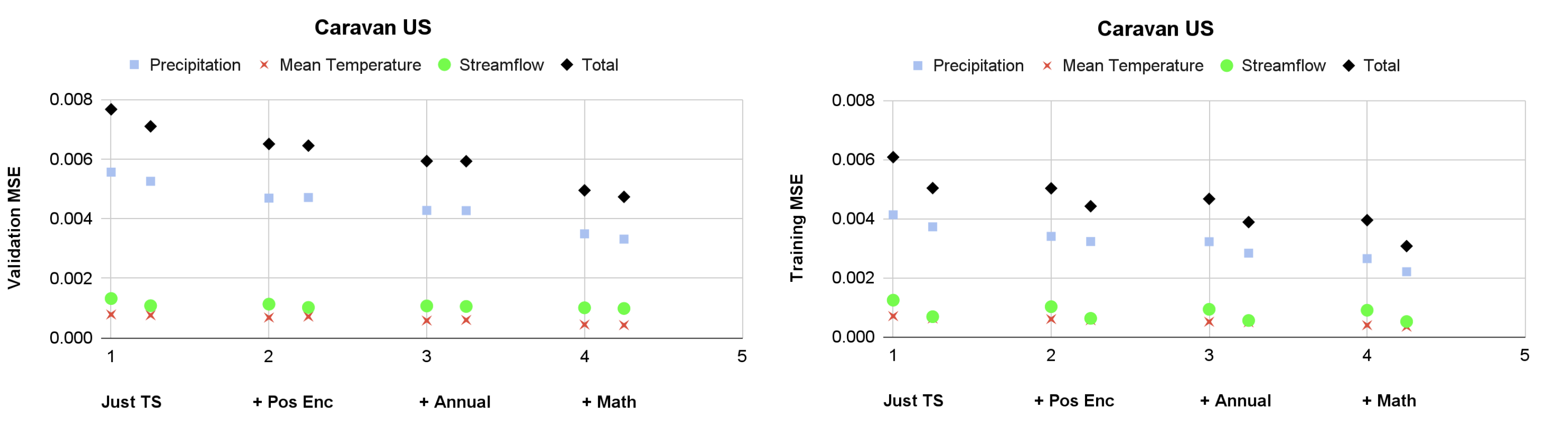}
\vspace{-4mm}
\caption{Caravan US experiment (RMSE loss). Each run configuration is utilized in 2 runs (left: static features not trained, right: static features trained).}
\label{caravan_us_static_encodings}
\vspace{3mm}

\includegraphics[width=\linewidth]{figs/CamelsCombined_results.png}
\vspace{-4mm}
\caption{Caravan combined experiment (RMSE loss). Each run configuration is utilized in 2 runs (left: static features not trained, right: static features trained).}
\label{caravan_combined_static_encodings}
\end{figure}

\subsection{Experiment Setup} 
This set of experiments examines the influence of static properties and spatial-temporal encodings on rainfall-runoff modeling accuracy, demonstrated through eight run configurations on four datasets. The first experiment uses the original CAMELS-US dataset \cite{camels-us-2017}, representing a traditional, locally-sourced small-scale dataset. Results are presented in Fig. \ref{camels_us_static_encodings}. The second experiment utilizes the Three-Nation Combined CAMELS dataset \cite{camels-us-2017, camels-gb-2020, camels-cl-2018}, representing a locally-sourced mid-scale dataset. Results are shown in Fig. \ref{camels_combined_static_encodings}. The third experiment is conducted on the US sub-dataset of the Caravan dataset \cite{caravan-2023}, representing a globally-sourced small-scale dataset. Results are shown in Fig. \ref{caravan_us_static_encodings}. The fourth experiment uses the full Caravan concatenated dataset \cite{caravan-2023}, representing a globally-sourced large-scale dataset. Results are shown in Fig. \ref{caravan_combined_static_encodings}. All four spatial-temporal encoding configurations tested are presented in Table \ref{camels_static_encoding_config}. For each configuration, we conduct runs with and without static features as input to highlight the impact of static features. The model architecture remained consistent across all runs.

\setlength{\tabcolsep}{2pt}
\renewcommand{\arraystretch}{1.4}
\begin{table}[htbp]
\caption{Static Properties and Encodings Experiment Configurations}
\begin{center}
\small
\begin{tabular}{ c c c c c }
\toprule
& \textbf{Config 1} & \textbf{Config 2} & \textbf{Config 3} & \textbf{Config 4} \\
\midrule
\textbf{Time Series Input}  & \checkmark    & \checkmark    & \checkmark    & \checkmark \\
\textbf{Linear Space}       &               & \checkmark    & \checkmark    & \checkmark \\
\textbf{Linear Time}        &               & \checkmark    & \checkmark    & \checkmark \\
\textbf{Annual Fourier Time}&               &               & \checkmark    & \checkmark \\
\textbf{Extra Fourier Time} &               &               &               & \checkmark \\
\textbf{Legendre Time}      &               &               &               & \checkmark \\
\bottomrule

\end{tabular}
\label{camels_static_encoding_config}
\end{center}
\vspace{-6mm}
\end{table}

\subsection{Experiment Findings} Results suggest that while the addition of static features in training has a marginal effect, the LSTM network generally benefits from their inclusion. This is demonstrated by the slightly lower training and validation losses in the runs conducted with static input features for each encoding configuration, as shown in Fig. \ref{camels_us_static_encodings}, Fig. \ref{camels_combined_static_encodings}, Fig. \ref{caravan_us_static_encodings}, and Fig. \ref{caravan_combined_static_encodings}. Furthermore, the figures indicate that static input features have a greater impact on large-scale datasets compared to small-scale ones, and they appear to be more beneficial for models trained on the Caravan dataset than those trained on the CAMELS datasets. We hypothesize that this is due to the greater number of static properties used in the Caravan runs—approximately five times more than in the CAMELS runs—and that larger datasets, covering a greater number of catchments, introduce more complexity.

This study further demonstrates that spatial and temporal encoding are crucial for effectively training time series data that follow seasonal patterns. The incorporation of linear spatial-temporal encoding, as evidenced by the drop in loss from run configuration 1 to run configuration 2 in the plots, captures both the spatial relationships of catchments and the time dependence of the data. The inclusion of annual Fourier temporal encoding, shown by the drop in loss from run configuration 2 to run configuration 3, captures the yearly seasonality of hydrological data. The addition of extra Fourier and Legendre temporal encoding, indicated by the drop in loss from run configuration 3 to run configuration 4, captures both known and unknown hydrological patterns of varying lengths, thereby enhancing model performance.

In hydrology, it is reasonable to assume that interrelated time series targets are highly dependent on static properties, such as land characteristics at individual gauges. Additionally, the water cycle is governed by processes with both known and unknown periodicities, ranging from atmospheric to micro scales. However, deep learning-based time series studies often overlook the significance of these characteristics, as they are typically application-specific. Notably, recent studies tend to focus on developing state-of-the-art time series forecasting models that rely solely on raw time series data. While these foundational models allow for broad applicability across various fields, they may compromise prediction accuracy in specialized downstream applications if static features or known patterns are not incorporated into the training process.

\section{Data Splitting and Cube Root Transformation Experiment}

Table \ref{tab:lstm_results} summarizes four experimental configurations for training and evaluating the LSTM model on precipitation, varying the train/validation set splitting (time-based vs.\ location-based) and preprocessing (just min-max scaling vs.\ min-max scaling and cube root transformation). The results, sorted by the Normalized Nash–Sutcliffe Efficiency (NNSE), shed light on how domain-specific considerations can significantly affect model performance. All runs from this experiment use the baseline LSTM from Nixtla’s NeuralForecast framework \cite{neuralforecast2022}, as the custom LSTM model introduced in this study currently supports only location-based splits.

\begin{table}[h!]
    \centering
    \caption{Performance of LSTM across different data preprocessing and splitting strategies.}
    \label{tab:lstm_results}
    \begin{tabular}{lccc}
        \hline
        \textbf{Splitting} & \textbf{Preprocessing} &  \textbf{RMSE} & \textbf{NNSE} \\
        \hline
        Time & min-max  & 0.0187 & 0.618 \\
        Time & min-max \& cube root& 0.0826 & 0.718 \\
        Location & min-max& 0.0141 & 0.744 \\
        Location & min-max \& cube root& 0.0604 & 0.824 \\
        \hline
    \end{tabular}
\end{table}

 Location-based splitting with cube root transformation achieves the highest NNSE (0.8240), demonstrating that LSTMs perform better in modeling rainfall-runoff relationships for unseen catchments when precipitation data are cube-root-transformed. This transformation reduces skewness caused by large variations in precipitation across catchments with differing climates and land surfaces, enabling better learning from extreme values. In comparison, using raw data with the same location-based split results in a lower NNSE (0.7443), likely due to skewed values disrupting stable training across diverse catchments.

Time-based and location-based splitting highlight different objectives in hydrological modeling. Time-based splitting aims at forecasting future runoff within the same catchments, making it suitable for operational predictions, but yields lower NNSE values (0.6177–0.7180). Location-based splitting evaluates a model’s ability to generalize to new catchments, which is essential for global-scale or unmonitored applications. The higher NNSE for location-based splitting suggests it better captures robust spatial patterns.

RMSE results, reported in Table \ref{tab:lstm_results}, vary across test sets (distinct years for time-based, distinct catchments for location-based) and are not directly comparable. NNSE, however, normalizes residuals by the observed variance, providing a consistent and reliable measure of model performance across different configurations. This makes NNSE the most appropriate metric for evaluating how well the model captures hydrological dynamics.

\begin{table*}[ht!]
\setlength{\tabcolsep}{5pt}
\setlength{\aboverulesep}{0pt}
\setlength{\belowrulesep}{0pt}
\centering
\caption{Performance comparison of various models}
\begin{tabular}{l|c|c|c|c|c|c|c|c}
\toprule
\textbf{Model} & 
\textbf{Type $^a$} &
\textbf{Input\&Output $^b$} &
\textbf{P $^c$} & 
\textbf{SR $^d$} & 
\textbf{Tmax $^e$} & 
\textbf{Tmin $^f$} & 
\textbf{VP $^g$} & 
\textbf{Q $^h$} \\
\toprule
LSTM-multivariate &
P &
MV\&MV &
\textcolor{red}{0.0546} &            
\textcolor{red}{0.0509} &            
\textcolor{red}{0.0138} &            
0.0197 &            
\textcolor{red}{0.0122} &            
\textcolor{red}{0.0161} \\           

LSTM-univariate &
P &
UV\&UV &
0.0583 &             
0.0567 &            
0.0195 &            
\textcolor{red}{0.0195} &            
0.0155 &            
0.0190 \\           

TFT &
P &
UV\&UV &
0.0953 &            
0.0918 &            
0.0277 &            
0.0305 &            
0.0257 &            
0.2385 \\           

TSMixer-M4 &
F &
MV\&MV &
0.0849 &            
0.0774 &            
0.0288 &            
0.0226 &            
0.0311 &            
0.0648 \\           

TSMixer-TrafficL &
F &
MV\&MV &
0.0859 &            
0.0780 &            
0.0293 &            
0.0230 &            
0.0318 &            
0.0210 \\           

TSMixer &
F &
MV\&MV &
0.0931 &            
0.0835 &            
0.0311 &            
0.0270 &            
0.0366 &            
0.0255 \\           

TCN &
P &
UV\&UV &
0.1036 &            
0.0943 &            
0.0344 &            
0.0329 &            
0.0383 &            
0.0535 \\           

iTransformer-M4 &
F &
MV\&MV &
0.0892 &            
0.0863 &            
0.0354 &            
0.0283 &            
0.0356 &            
0.0202 \\           

iTransformer &
F &
MV\&MV &
0.0869 &            
0.0881 &            
0.0356 &            
0.0257 &            
0.0351 &            
0.0232 \\           

DilatedRNN &
P &
UV\&UV &
0.0993 &            
0.0947 &            
0.0358 &            
0.0313 &            
0.0375 &            
0.0279 \\           

iTransformer-TrafficL &
F &
MV\&MV &
0.0920 &            
0.0889 &            
0.0374 &            
0.0307 &            
0.0373 &            
0.0205 \\           

GNNCoder-univariate &
P &
UV\&UV &
0.0843 &          
0.0884 &           
0.0389 &           
0.0345 &           
0.0395 &           
0.0200 \\          

PatchTST-Weather &
F &
UV\&UV &
0.0944 &            
0.0920 &            
0.0411 &            
0.0335 &            
0.0405 &            
0.0207 \\           

PatchTST-TrafficL &
F &
UV\&UV &
0.0943 &            
0.0920 &            
0.0411 &            
0.0333 &            
0.0406 &            
0.0207 \\           

PatchTST-M4 &
F &
UV\&UV &
0.0944 &            
0.0920 &            
0.0411 &            
0.0333 &            
0.0407 &            
0.0207 \\           

GNNCoder-multivariate &
P &
MV\&MV &
0.0928 &           
0.0407 &           
0.0414 &           
0.0893 &           
0.0497 &        
0.0200 \\         

TimesNet &
P &
UV\&UV &
0.0969 &            
0.0934 &            
0.0420 &            
0.0345 &            
0.0421 &            
0.2943 \\           

PatchTST &
F &
UV\&UV &
0.0976 &            
0.0932 &            
0.0429 &            
0.0335 &            
0.0409 &            
0.0235 \\           

TiDE &
P &
UV\&UV &
0.1014 &            
0.0933 &            
0.0452 &            
0.0400 &            
0.0518 &            
0.0310 \\           

Chronos-T5 &
F &
UV\&UV &
0.1895 &             
0.1236 &            
0.0792 &            
0.0753 &            
0.0851 &            
0.1403 \\           

Chronos-GPT2 &
F &
UV\&UV &
0.1675 &            
0.1123 &            
0.0794 &            
0.0814 &            
0.1601 &            
0.1193 \\           

\bottomrule

\multicolumn{9}{l}{\small $^a$Pattern or foundation model, $^b$UniVariate or MultiVariate training and forecasting} \\
\multicolumn{9}{l}{\small $^c$Precipitation, $^d$Solar radiation, $^e$Maximum temperature, $^f$Minimum temperature, $^g$Vapor pressure, $^h$Streamflow.} \\
\multicolumn{9}{l}{\small Table shows root mean squared error results, with the lowest for each predicted target highlighted in red.} \\
\end{tabular}
\label{foundation_model_nixtla}
\end{table*}

\section{Comparison with other Pattern and Foundation Models}

Recent advancements in time series analysis have adopted a foundation model approach, drawing inspiration from large language models \cite{mirchandani2023largelanguagemodelsgeneral}. Since 2022, there has been a surge in research on foundation models for time series forecasting, primarily utilizing Transformer \cite{vaswani2023attentionneed} and MLP \cite{tolstikhin2021mlpmixerallmlparchitecturevision} architectures. Some models incorporate static exogenous variables as input \cite{chen2023tsmixerallmlparchitecturetime, das2023long}, while others rely solely on time series data \cite{ansari2024chronoslearninglanguagetime}. These models aim to generalize well in data-scarce tasks, integrate multimodal data, and offer explainable decision-making processes \cite{ye2024surveytimeseriesfoundation}. Research in this area focuses on two strategies: pre-training foundation models with time series data and adapting pre-trained language models for time series tasks \cite{ye2024surveytimeseriesfoundation}. Their success stems from effectively capturing multimodal distributions and repeated seasonal patterns \cite{gruver2024large}.

This study compares the performance of state-of-the-art foundation models and our LSTM-based pattern model on the hydrology rainfall-runoff task. Pattern models, smaller-scale models trained for specific tasks \cite{Jafari2024Earthquake}, are contrasted with broad-purpose foundation models. All benchmarks use CAMELS-US \cite{camels-us-2017} data and the Nixtla NeuralForecast framework \cite{neuralforecast2022}. Pattern models include TFT \cite{lim2021temporal}, TCN \cite{bai2018empirical}, DilatedRNN \cite{chang2017dilated}, GNN-Coder \cite{Jafari2024Earthquake}, TimesNet \cite{wu2022timesnet}, and TiDE \cite{das2023long}. Foundation models analyzed include TSMixer \cite{chen2023tsmixerallmlparchitecturetime}, iTransformer \cite{liu2023itransformer}, and PatchTST \cite{nie2022time}, pre-trained on datasets like TrafficL \cite{lai2018modeling}, M4 \cite{makridakis2020m4}, and Weather \cite{wu2021autoformer}. Adapted models from LLMs, such as Chronos \cite{ansari2024chronoslearninglanguagetime}, are also evaluated, with Chronos using T5 \cite{JMLR:v21:20-074} and GPT-2 \cite{radford2019language} backbones. For all models, (pattern and foundation) the backbone components are unfrozen during training and fine-tuning. This choice is informed by empirical observations that freezing model backbones leads to a slight drop in performance on CAMELS-US data.

Results are presented through two experiments: one using min-max scaling and cube root transformation and the other using raw precipitation data.

\begin{figure}[htbp]
\centering
\includegraphics[width=\linewidth]{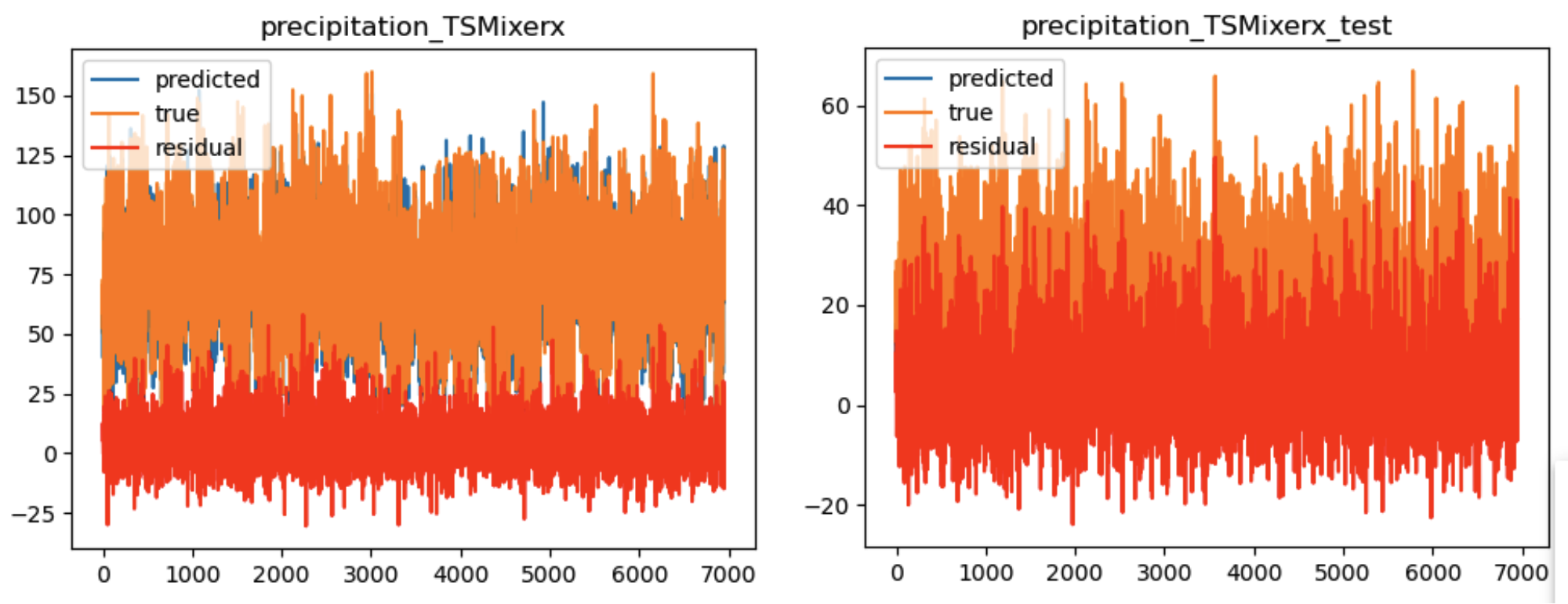}
\caption{TSMixer precipitation fit (left: training set, right: validation set).}
\label{tsmixer_prcp}
\vspace{3mm}

\includegraphics[width=\linewidth]{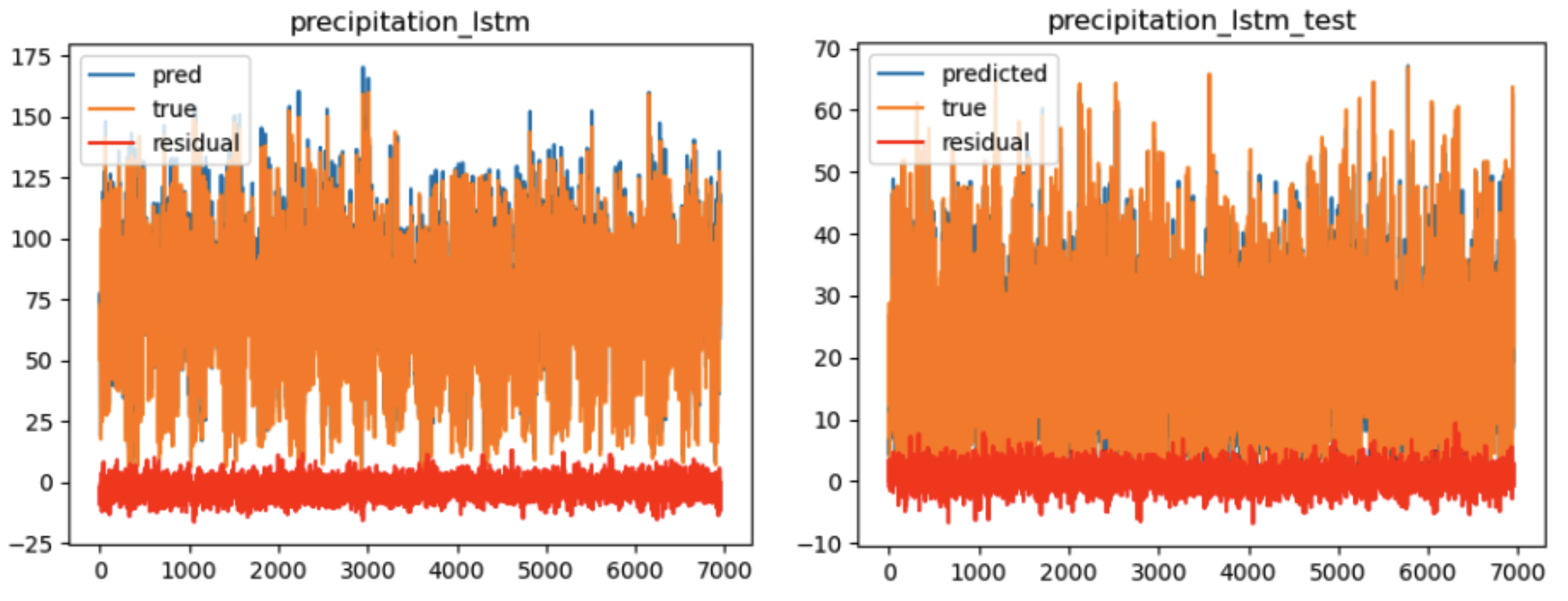}
\caption{LSTM precipitation fit (left: training set, right: validation set).}
\label{lstm_prcp}
\vspace{3mm}

\includegraphics[width=\linewidth]{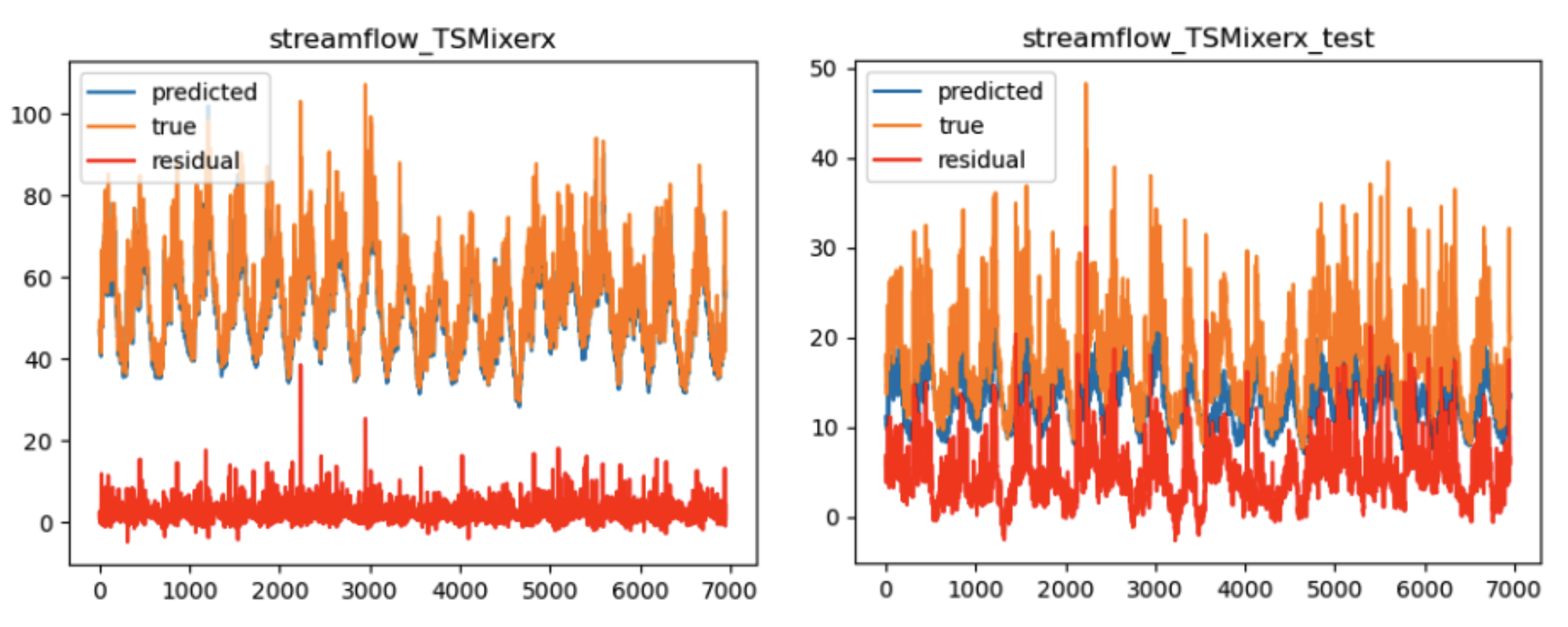}
\caption{TSMixer streamflow fit (left: training set, right: validation set).}
\label{tsmixer_strmflow}
\vspace{3mm}

\includegraphics[width=\linewidth]{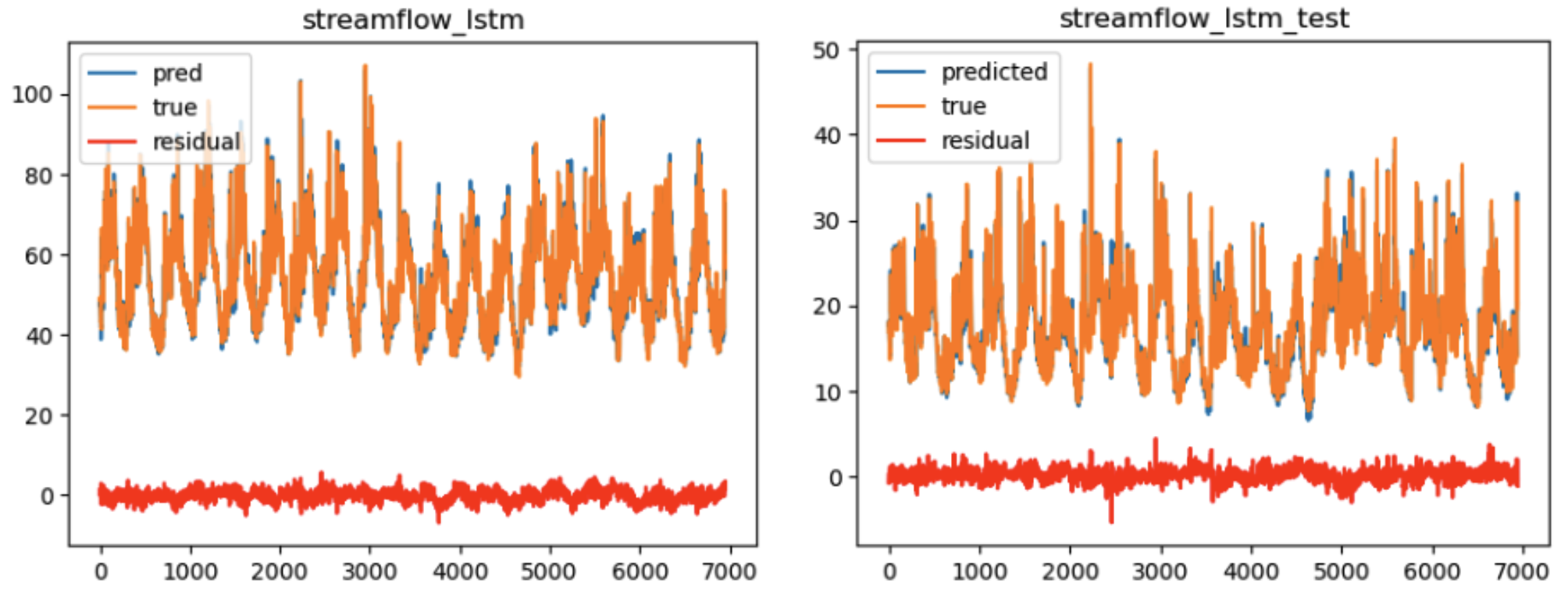}
\caption{LSTM streamflow fit (left: training set, right: validation set).}
\label{lstm_strmflow}
\end{figure}

\begin{figure}[htbp]
\centering
\includegraphics[width=\linewidth]{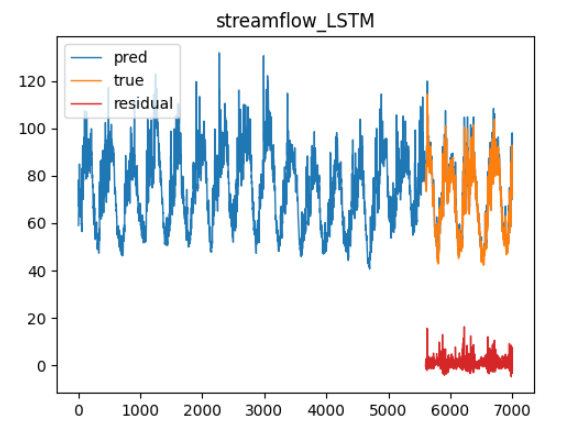}
\caption{LSTM streamflow fit (temporal validation).}
\label{lstm_time}
\vspace{3mm}

\includegraphics[width=\linewidth]{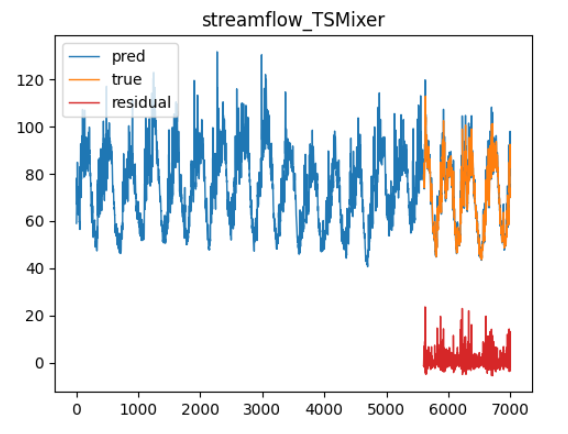}
\caption{TSMixer streamflow fit (temporal validation).}
\label{tsmixer_time}
\vspace{3mm}

\includegraphics[width=\linewidth]{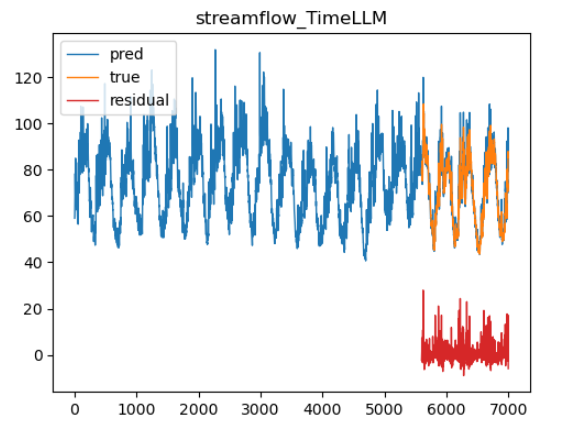}
\caption{TimeLLM streamflow fit (temporal validation).}
\label{timellm_time}
\vspace{3mm}
\end{figure}

\begin{table*}[ht!]
    \centering
    \setlength{\tabcolsep}{5pt}
    \caption{Performance of time-segmented precipitation forecasting models, where “P” shows pattern models, “F” indicates foundation models, “MV” stands for multivariate inputs/outputs, and “UV” for univariate.}
    \label{tab:raw_results}
    \begin{tabular}{l|c|c|c|c}
        \hline
        \textbf{Model} & \textbf{Type} & \textbf{Input\&Output} & \textbf{RMSE} & \textbf{NNSE} \\
        \hline
        TSMixer-M4 & F & UV\&UV & 5.9916 & 0.5623 \\
        iTransformer-M4 & F & UV\&UV & 5.7462 & 0.5827 \\
        iTransformer-TrafficL & F & UV\&UV & 5.7288 & 0.5842 \\
        TSMixer-TrafficL & F & UV\&UV & 5.6505 & 0.5909 \\
        PatchTST-Weather & F & UV\&UV & 5.6001 & 0.5952 \\
        TimesNet & P & MV\&UV & 5.5493 & 0.5996 \\
        TimeLLM & F & UV\&UV & 5.5208 & 0.6021 \\
        TCN & P & MV\&UV & 5.4976 & 0.6041 \\
        PatchTST-TrafficL & F & UV\&UV & 5.4665 & 0.6068 \\
        TiDE & P & MV\&UV & 5.4631 & 0.6071 \\
        PatchTST-M4 & F & UV\&UV & 5.4347 & 0.6096 \\
        DilatedRNN & P & MV\&UV & 5.2325 & 0.6275 \\
        TFT & P & MV\&UV & 5.2110 & 0.6294 \\
        LSTM & P & MV\&UV & 5.2022 & 0.6302 \\
        MultiFoundationPattern & F & MV\&UV & 5.1921 & 0.6305 \\
        MultiFoundationCore & F & MV\&UV & \textcolor{red}{5.1802} & \textcolor{red}{0.6312} \\
        \hline
    \end{tabular}
\end{table*}

\subsection{Model Performance Comparison of Processed Data Forecasting (Spatial Validation)}

Our LSTM model is trained under benchmark configurations in both univariate and multivariate settings, using all 6 time series properties. Missing streamflow values are imputed with the mean of all observed streamflow data for both univariate and multivariate training. We apply min-max scaling and cube root transformation on precipitation and streamflow data. All foundation and pattern models are trained and evaluated on the same training and validation locations, with a sequence length of 21 days and a forecast horizon of 1 day. Static exogenous features are utilized if supported by the model. As shown by the root mean squared error results in Table \ref{foundation_model_nixtla}, the multivariately trained LSTM model outperforms all other approaches under these benchmark conditions. Note that this study did not study the best hyperparameters that yield the best results for each model.

Model fits shown in Fig.~\ref{tsmixer_prcp}, Fig.~\ref{lstm_prcp}, Fig.~\ref{tsmixer_strmflow}, and Fig.~\ref{lstm_strmflow} illustrate LSTM model performance compared to the TSMixer model \cite{chen2023tsmixerallmlparchitecturetime} that allows static exogenous features along with multivariate time series as input.

Most foundation models prioritize univariate forecasting for broad applicability \cite{ansari2024chronoslearninglanguagetime}. However, a multivariate approach, which jointly trains and forecasts correlated time series properties, can improve performance. Using CAMELS-US data \cite{camels-us-2017}, we found that multivariate training outperformed univariate methods, particularly in tasks like rainfall-runoff modeling where correlations between properties are crucial.

\subsection{Model Performance Comparison of Raw Precipitation Forecasting (Temporal Validation)}

The evaluation of raw precipitation data reveals notable performance differences among the tested models, as summarized in Table \ref{tab:raw_results} (RMSE and NNSE).

The MultiFoundationCore achieved the best results (RMSE: 5.1802, NNSE: 0.6312), closely followed by the MultiFoundationPattern (RMSE: 5.1921, NNSE: 0.6305). These advanced architectures effectively model temporal patterns in precipitation, making them robust solutions for hydrological applications.

Among baseline models, LSTM and TFT delivered strong results, with LSTM achieving an NNSE of 0.6302 (RMSE: 5.2022) and TFT slightly behind with an NNSE of 0.6294 (RMSE: 5.2110). Other models, such as TimeLLM  (NNSE: 0.6021) and TCN (NNSE: 0.6041), captured temporal dependencies effectively but lagged behind the more advanced architectures.

Foundation models like TSMixer-M4 (RMSE: 5.9916) and iTransformer-M4 (RMSE: 5.7462) struggled to adapt to raw precipitation data, underscoring the need for domain-specific adjustments. Hybrid models such as DilatedRNN performed competitively (RMSE: 5.2325), while models like PatchTST-M4 and TiDE showed moderate results, with NNSE values of 0.6096 and 0.6071, respectively, indicating limitations in capturing the complexities of raw precipitation data.

Model fits shown in Fig.~\ref{lstm_time}, Fig.~\ref{tsmixer_time}, and Fig.~\ref{timellm_time} illustrate LSTM model performance compared to the TSMixer model \cite{chen2023tsmixerallmlparchitecturetime} and TimeLLM model \cite{jin2024timellmtimeseriesforecasting} with temporal validation.

The MultiFoundationCore architecture demonstrates strong adaptability and scalability in hydrology, effectively capturing intricate patterns across various preprocessing techniques and data-splitting strategies. Its robust performance underscores its suitability for diverse hydrological applications

\section{MultiFoundationCore Model}

The MultiFoundationCore framework enhances time series forecasting in complex scientific domains by integrating multiple pre-trained foundation models and specialized pattern models. Building on the MultiFoundationPattern \cite{Jafari2024Earthquake}, it uses advanced cross-attention mechanisms to synthesize diverse inputs, capturing intricate spatial and temporal dependencies. This scalable approach leverages the strengths of various models to improve prediction accuracy.

Initial results in the hydrology domain show significant improvements in forecasting accuracy, particularly for complex spatiotemporal patterns. These findings underscore the framework's potential to generalize across scientific applications, advancing reliable time series forecasting in challenging domains.

\section{Conclusion and Future Work}
Often, one thinks about deep learning models discovering hidden variables that control observables that one wishes to predict. In this paper, we demonstrate how including environmental factors and exogenous features can improve time series models' ability to learn hidden variables. We explored CAMELS and Caravan, which together offer large datasets covering many (about 8000) spatial locations and a long time period (around 30 years). Further, there are multiple (6 in CAMELS and 39 in Caravan) observed dynamic streams and up to 209 static (exogenous) parameters for each location. Some of these exogenous and observed data would be natural inputs into a physics model for the catchments. A priori, it is not obvious which exogenous or even endogenous data should be used as they could be implied by the large datasets used in the training and implicitly learned by the training. However, most likely, it is a mixed situation where the combination of observed and exogenous data will give the best results, with the exogenous data declining in importance as the observed dataset grows in size. We clearly show in this paper with the CAMELS and Caravan data that using exogenous data makes significant improvements in the model fits where both static physical quantities and dynamical mathematical functions have value as exogenous information. We emphasize the importance of considering all available data in time series analyses. In future papers, we hope to understand the trade-off better as it varies in the nature and size of exogenous data. We also will present in-depth study of time series pattern and foundation models under varying hyperparameters, using the compilation of hundreds of papers and around 100 models in \cite{sciencefmhub, nixtla, ddz, survey2023, survey2024}. We aim to study a range of time series problems to build a taxonomy so that benchmark sets can be built covering the essentially different cases \cite{Huang2019mlc}. Through a comprehensive model review, we will also examine univariate versus multivariate time series analysis. Another interesting feature of science time series is that they naturally vary in magnitude by large factors, and this seems not to be very consistent with neural network activation functions at fixed values independent of the stream. We will present a study of new activation functions in the LSTM using a PRELU-like (Parametric Rectified Linear Unit, or PReLU) structure, which naturally deals with magnitude changes in stream values.

Our LSTM model, data preprocessing, and model comparison code is available at \url{https://github.com/JunyangHe/Hydrology}. The processed data is available at \cite{he_2024_13975174}.

\section*{Acknowledgment}

We would like to thank Gregor von Laszewski and Niranda Perera for their contributions and guidance. We gratefully acknowledge the partial support of DE-SC0023452: FAIR Surrogate Benchmarks Supporting AI and Simulation Research and the Biocomplexity Institute at the University of Virginia.

\bibstyle{IEEEtran}
\bibliography{references}

\begin{thebibliography}{10}
\providecommand{\url}[1]{#1}
\csname url@samestyle\endcsname
\providecommand{\newblock}{\relax}
\providecommand{\bibinfo}[2]{#2}
\providecommand{\BIBentrySTDinterwordspacing}{\spaceskip=0pt\relax}
\providecommand{\BIBentryALTinterwordstretchfactor}{4}
\providecommand{\BIBentryALTinterwordspacing}{\spaceskip=\fontdimen2\font plus
\BIBentryALTinterwordstretchfactor\fontdimen3\font minus \fontdimen4\font\relax}
\providecommand{\BIBforeignlanguage}[2]{{%
\expandafter\ifx\csname l@#1\endcsname\relax
\typeout{** WARNING: IEEEtran.bst: No hyphenation pattern has been}%
\typeout{** loaded for the language `#1'. Using the pattern for}%
\typeout{** the default language instead.}%
\else
\language=\csname l@#1\endcsname
\fi
#2}}
\providecommand{\BIBdecl}{\relax}
\BIBdecl

\bibitem{cdcNationalCenter}
CDC, ``{N}ational {C}enter for {H}ealth {S}tatistics --- cdc.gov,'' \url{https://www.cdc.gov/nchs/index.html}, n.d., [Accessed 04-10-2024].

\bibitem{cdcBehavioralRisk}
``{B}ehavioral {R}isk {F}actor {S}urveillance {S}ystem --- cdc.gov,'' \url{https://www.cdc.gov/brfss/index.html}, n.d., [Accessed 04-10-2024].

\bibitem{covid-2021}
G.~C. Fox, G.~von Laszewski, F.~Wang, and S.~Pyne, ``Aicov: An integrative deep learning framework for covid-19 forecasting with population covariates,'' \emph{Journal of Data Science}, vol.~19, no.~2, pp. 293--313, 2021.

\bibitem{usgsSearchEarthquake}
``Earthquake hazards program of united states geological survey,'' \url{https://earthquake.usgs.gov/earthquakes/search/}, n.d., [Accessed 04-10-2024].

\bibitem{earthquake-2022}
\BIBentryALTinterwordspacing
G.~C. Fox, J.~B. Rundle, A.~Donnellan, and B.~Feng, ``Earthquake nowcasting with deep learning,'' \emph{GeoHazards}, vol.~3, no.~2, pp. 199--226, 2022. [Online]. Available: \url{https://www.mdpi.com/2624-795X/3/2/11}
\BIBentrySTDinterwordspacing

\bibitem{sciencefmhub}
``{ScienceFMHub} portal for science foundation model community,'' \url{http://sciencefmhub.org}, 2~Nov. 2023, accessed: 2023-11-3.

\bibitem{nixtla}
``Nixtla: A future for everybody. time series research and deployment,'' \url{https://www.nixtla.io/}, n.d., accessed: 2024-3-9.

\bibitem{ddz}
{ddz}, ``Awesome time series forecasting/prediction papers: a reading list of papers on time series forecasting/prediction ({TSF}) and spatio-temporal forecasting/prediction ({STF}),'' \url{https://github.com/ddz16/TSFpaper}, n.d., accessed: 2024-5-16.

\bibitem{survey2023}
\BIBentryALTinterwordspacing
M.~Jin, Q.~Wen, Y.~Liang, C.~Zhang, S.~Xue, X.~Wang, J.~Zhang, Y.~Wang, H.~Chen, X.~Li, S.~Pan, V.~S. Tseng, Y.~Zheng, L.~Chen, and H.~Xiong, ``Large models for time series and spatio-temporal data: A survey and outlook,'' \emph{arXiv [cs.LG]}, 16~Oct. 2023. [Online]. Available: \url{http://arxiv.org/abs/2310.10196}
\BIBentrySTDinterwordspacing

\bibitem{survey2024}
\BIBentryALTinterwordspacing
Y.~Liang, H.~Wen, Y.~Nie, Y.~Jiang, M.~Jin, D.~Song, S.~Pan, and Q.~Wen, ``Foundation models for time series analysis: A tutorial and survey,'' \emph{arXiv [cs.LG]}, 21~Mar. 2024. [Online]. Available: \url{http://arxiv.org/abs/2403.14735}
\BIBentrySTDinterwordspacing

\bibitem{Jafari2024Earthquake}
\BIBentryALTinterwordspacing
A.~Jafari, G.~Fox, J.~B. Rundle, A.~Donnellan, and L.~G. Ludwig, ``Time series foundation models and deep learning architectures for earthquake temporal and spatial nowcasting,'' \emph{arXiv [cs.LG, physics.geo-ph]}, 22~Aug. 2024. [Online]. Available: \url{http://arxiv.org/abs/2408.11990}
\BIBentrySTDinterwordspacing

\bibitem{dingman2015physical}
S.~L. Dingman, \emph{Physical hydrology}.\hskip 1em plus 0.5em minus 0.4em\relax Waveland press, 2015.

\bibitem{mulvaney_1851}
\BIBentryALTinterwordspacing
M.~T. J., ``On the use of self-registering rain and flood gauges in making observations of the relations of rainfall and flood discharges in a given catchment,'' \emph{Proceedings of the Institution of Civil Engineers of Ireland}, vol.~4, pp. 19--31, 1851. [Online]. Available: \url{https://cir.nii.ac.jp/crid/1573105975719082240}
\BIBentrySTDinterwordspacing

\bibitem{Solr-29746}
N.~H. Crawford, \emph{Digital simulation in hydrology : Stanford watershed model IV / by N.H. Crawford and Ray K. Linsley.}, ser. Stanford University Department of Civil Engineering Technical report ;, .~Linsley, Ray K. (Ray~Keyes), Ed.\hskip 1em plus 0.5em minus 0.4em\relax Stanford University. Dept of Civil Engineering, 1966.

\bibitem{beven_rainfall-runoff_2012}
K.~J. Beven, {Ebook Central - Academic Complete}, {Wiley Online Library UBCM Earth \& Environmental}, {Wiley Online Library UBCM German Language}, and {Wiley Online Library UBCM All Obooks}, \emph{\BIBforeignlanguage{English}{Rainfall-runoff {Modelling}: {The} {Primer}}}.\hskip 1em plus 0.5em minus 0.4em\relax Chichester, West Sussex, Hoboken, NJ: Wiley-Blackwell, 2012.

\bibitem{ABBOTT198645}
\BIBentryALTinterwordspacing
M.~Abbott, J.~Bathurst, J.~Cunge, P.~O'Connell, and J.~Rasmussen, ``An introduction to the european hydrological system — systeme hydrologique europeen, “she”, 1: History and philosophy of a physically-based, distributed modelling system,'' \emph{Journal of Hydrology}, vol.~87, no.~1, pp. 45--59, 1986. [Online]. Available: \url{https://www.sciencedirect.com/science/article/pii/0022169486901149}
\BIBentrySTDinterwordspacing

\bibitem{moore_clark_pdm_1981}
R.~J. Moore and R.~T. Clarke, ``A distribution function approach to rainfall runoff modeling,'' \emph{Water Resources Research}, vol.~17, no.~5, pp. 1367--1382, 1981.

\bibitem{overview_rainfall_runoff_2017}
J.~Sitterson, R.~P. Chris~Knightes, K.~Wolfe, M.~Muche, and B.~Avant, \emph{An Overview of Rainfall-Runoff Model Types}.\hskip 1em plus 0.5em minus 0.4em\relax U.S. Environmental Protection Agency, Washington, DC, 2017.

\bibitem{kratzert-lstm-hydrology-2018}
\BIBentryALTinterwordspacing
F.~Kratzert, D.~Klotz, C.~Brenner, K.~Schulz, and M.~Herrnegger, ``Rainfall--runoff modelling using long short-term memory (lstm) networks,'' \emph{Hydrology and Earth System Sciences}, vol.~22, no.~11, pp. 6005--6022, 2018. [Online]. Available: \url{https://hess.copernicus.org/articles/22/6005/2018/}
\BIBentrySTDinterwordspacing

\bibitem{shen_2018}
\BIBentryALTinterwordspacing
C.~Shen, ``A transdisciplinary review of deep learning research and its relevance for water resources scientists,'' \emph{Water Resources Research}, vol.~54, no.~11, pp. 8558--8593, 2018. [Online]. Available: \url{https://agupubs.onlinelibrary.wiley.com/doi/abs/10.1029/2018WR022643}
\BIBentrySTDinterwordspacing

\bibitem{hsu_artificial_neural_network_1995}
\BIBentryALTinterwordspacing
K.-l. Hsu, H.~V. Gupta, and S.~Sorooshian, ``Artificial neural network modeling of the rainfall-runoff process,'' \emph{Water Resources Research}, vol.~31, no.~10, pp. 2517--2530, 1995. [Online]. Available: \url{https://agupubs.onlinelibrary.wiley.com/doi/abs/10.1029/95WR01955}
\BIBentrySTDinterwordspacing

\bibitem{tokar_a_sezin_rainfall-runoff_1999}
\BIBentryALTinterwordspacing
{Tokar A. Sezin} and {Johnson Peggy A.}, ``Rainfall-{Runoff} {Modeling} {Using} {Artificial} {Neural} {Networks},'' \emph{Journal of Hydrologic Engineering}, vol.~4, no.~3, pp. 232--239, Jul. 1999, publisher: American Society of Civil Engineers. [Online]. Available: \url{https://doi.org/10.1061/(ASCE)1084-0699(1999)4:3(232)}
\BIBentrySTDinterwordspacing

\bibitem{ann-book-2013}
R.~S. Govindaraju and A.~R. Rao, \emph{\BIBforeignlanguage{en}{Artificial Neural Networks in Hydrology}}.\hskip 1em plus 0.5em minus 0.4em\relax Springer Science \& Business Media, Mar. 2013.

\bibitem{hu-lstm-hydrology-2018}
\BIBentryALTinterwordspacing
C.~Hu, Q.~Wu, H.~Li, S.~Jian, N.~Li, and Z.~Lou, ``Deep learning with a long short-term memory networks approach for rainfall-runoff simulation,'' \emph{Water}, vol.~10, no.~11, 2018. [Online]. Available: \url{https://www.mdpi.com/2073-4441/10/11/1543}
\BIBentrySTDinterwordspacing

\bibitem{Kratzert2019}
\BIBentryALTinterwordspacing
F.~Kratzert, M.~Herrnegger, D.~Klotz, S.~Hochreiter, and G.~Klambauer, \emph{NeuralHydrology -- Interpreting LSTMs in Hydrology}.\hskip 1em plus 0.5em minus 0.4em\relax Cham: Springer International Publishing, 2019, pp. 347--362. [Online]. Available: \url{https://doi.org/10.1007/978-3-030-28954-6\_19}
\BIBentrySTDinterwordspacing

\bibitem{kratzert_neuralhydrology_2022}
\BIBentryALTinterwordspacing
F.~Kratzert, M.~Gauch, G.~Nearing, and D.~Klotz, ``\BIBforeignlanguage{en}{{NeuralHydrology} --- {A} {Python} library for {Deep} {Learning} research in hydrology},'' \emph{\BIBforeignlanguage{en}{Journal of Open Source Software}}, vol.~7, no.~71, p. 4050, Mar. 2022. [Online]. Available: \url{https://joss.theoj.org/papers/10.21105/joss.04050}
\BIBentrySTDinterwordspacing

\bibitem{camels-us-2017}
\BIBentryALTinterwordspacing
N.~Addor, A.~J. Newman, N.~Mizukami, and M.~P. Clark, ``The camels data set: catchment attributes and meteorology for large-sample studies,'' \emph{Hydrology and Earth System Sciences}, vol.~21, no.~10, pp. 5293--5313, 2017. [Online]. Available: \url{https://hess.copernicus.org/articles/21/5293/2017/}
\BIBentrySTDinterwordspacing

\bibitem{camels-gb-2020}
\BIBentryALTinterwordspacing
G.~Coxon, N.~Addor, J.~P. Bloomfield, J.~Freer, M.~Fry, J.~Hannaford, N.~J.~K. Howden, R.~Lane, M.~Lewis, E.~L. Robinson, T.~Wagener, and R.~Woods, ``Camels-gb: hydrometeorological time series and landscape attributes for 671 catchments in great britain,'' \emph{Earth System Science Data}, vol.~12, no.~4, pp. 2459--2483, 2020. [Online]. Available: \url{https://essd.copernicus.org/articles/12/2459/2020/}
\BIBentrySTDinterwordspacing

\bibitem{camels-cl-2018}
\BIBentryALTinterwordspacing
C.~Alvarez-Garreton, P.~A. Mendoza, J.~P. Boisier, N.~Addor, M.~Galleguillos, M.~Zambrano-Bigiarini, A.~Lara, C.~Puelma, G.~Cortes, R.~Garreaud, J.~McPhee, and A.~Ayala, ``The camels-cl dataset: catchment attributes and meteorology for large sample studies -- chile dataset,'' \emph{Hydrology and Earth System Sciences}, vol.~22, no.~11, pp. 5817--5846, 2018. [Online]. Available: \url{https://hess.copernicus.org/articles/22/5817/2018/}
\BIBentrySTDinterwordspacing

\bibitem{camels-aus-2021}
\BIBentryALTinterwordspacing
K.~J.~A. Fowler, S.~C. Acharya, N.~Addor, C.~Chou, and M.~C. Peel, ``Camels-aus: hydrometeorological time series and landscape attributes for 222 catchments in australia,'' \emph{Earth System Science Data}, vol.~13, no.~8, pp. 3847--3867, 2021. [Online]. Available: \url{https://essd.copernicus.org/articles/13/3847/2021/}
\BIBentrySTDinterwordspacing

\bibitem{camels-br-2020}
\BIBentryALTinterwordspacing
V.~B.~P. Chagas, P.~L.~B. Chaffe, N.~Addor, F.~M. Fan, A.~S. Fleischmann, R.~C.~D. Paiva, and V.~A. Siqueira, ``Camels-br: hydrometeorological time series and landscape attributes for 897 catchments in brazil,'' \emph{Earth System Science Data}, vol.~12, no.~3, pp. 2075--2096, 2020. [Online]. Available: \url{https://essd.copernicus.org/articles/12/2075/2020/}
\BIBentrySTDinterwordspacing

\bibitem{camels-ch-2023}
\BIBentryALTinterwordspacing
M.~H\"oge, M.~Kauzlaric, R.~Siber, U.~Sch\"onenberger, P.~Horton, J.~Schwanbeck, M.~G. Floriancic, D.~Viviroli, S.~Wilhelm, A.~E. Sikorska-Senoner, N.~Addor, M.~Brunner, S.~Pool, M.~Zappa, and F.~Fenicia, ``Camels-ch: hydro-meteorological time series and landscape attributes for 331 catchments in hydrologic switzerland,'' \emph{Earth System Science Data}, vol.~15, no.~12, pp. 5755--5784, 2023. [Online]. Available: \url{https://essd.copernicus.org/articles/15/5755/2023/}
\BIBentrySTDinterwordspacing

\bibitem{camels-se-2024}
C.~Teutschbein, ``Camels‐se : Long‐term hydroclimatic observations (1961–2020) across 50 catchments in sweden as a resource for modelling, education, and collaboration,'' \emph{Geoscience Data Journal}, 2024.

\bibitem{camels-fr-2022}
\BIBentryALTinterwordspacing
O.~Delaigue, P.~Brigode, V.~Andr{\'e}assian, C.~Perrin, P.~Etchevers, J.-M. Soubeyroux, B.~Janet, and N.~Addor, ``Camels-fr: A large sample hydroclimatic dataset for france to explore hydrological diversity and support model benchmarking,'' in \emph{IAHS-2022 Scientific Assembly}, Montpellier, France, May 2022. [Online]. Available: \url{https://hal.inrae.fr/hal-03687235}
\BIBentrySTDinterwordspacing

\bibitem{maurer-2013}
\BIBentryALTinterwordspacing
B.~Livneh, E.~A. Rosenberg, C.~Lin, B.~Nijssen, V.~Mishra, K.~M. Andreadis, E.~P. Maurer, and D.~P. Lettenmaier, ``A long-term hydrologically based dataset of land surface fluxes and states for the conterminous united states: Update and extensions,'' \emph{Journal of Climate}, vol.~26, no.~23, pp. 9384 -- 9392, 2013. [Online]. Available: \url{https://journals.ametsoc.org/view/journals/clim/26/23/jcli-d-12-00508.1.xml}
\BIBentrySTDinterwordspacing

\bibitem{usgs-2012}
\BIBentryALTinterwordspacing
H.~Lins, ``"usgs hydro-climatic data network 2009 (hcdn–2009)",'' \emph{"U.S. Geological Survey Fact Sheet 2012–3047"}, 2012. [Online]. Available: \url{"https://pubs.usgs.gov/fs/2012/3047/"}
\BIBentrySTDinterwordspacing

\bibitem{ceh-gear-2015}
\BIBentryALTinterwordspacing
V.~D.~J. Keller, M.~Tanguy, I.~Prosdocimi, J.~A. Terry, O.~Hitt, S.~J. Cole, M.~Fry, D.~G. Morris, and H.~Dixon, ``Ceh-gear: 1 km resolution daily and monthly areal rainfall estimates for the uk for hydrological and other applications,'' \emph{Earth System Science Data}, vol.~7, no.~1, pp. 143--155, 2015. [Online]. Available: \url{https://essd.copernicus.org/articles/7/143/2015/}
\BIBentrySTDinterwordspacing

\bibitem{nrfa-2004}
J.~Hannaford, ``Development of a strategic data management system for a national hydrological database, the uk national river flow archive,'' \emph{Hydroinformatics}, pp. 637--644, 2004.

\bibitem{chess-met-2017}
\BIBentryALTinterwordspacing
E.~Robinson, E.~Blyth, D.~Clark, E.~Comyn-Platt, J.~Finch, and A.~Rudd, ``Climate hydrology and ecology research support system meteorology dataset for great britain (1961-2015) [chess-met] v1.2,'' 2017. [Online]. Available: \url{https://doi.org/10.5285/b745e7b1-626c-4ccc-ac27-56582e77b900}
\BIBentrySTDinterwordspacing

\bibitem{crmet-2023}
J.~P. Boisier, ``Cr2met: A high-resolution precipitation and temperature dataset for the period 1960-2021 in continental chile.'' Jan. 2023.

\bibitem{cr-streamflow}
\BIBentryALTinterwordspacing
CR2, ``Explorador climÁtico,'' 2016. [Online]. Available: \url{https://explorador.cr2.cl/}
\BIBentrySTDinterwordspacing

\bibitem{caravan-2023}
\BIBentryALTinterwordspacing
F.~Kratzert, G.~Nearing, N.~Addor, T.~Erickson, M.~Gauch, O.~Gilon, L.~Gudmundsson, A.~Hassidim, D.~Klotz, S.~Nevo, G.~Shalev, and Y.~Matias, ``Caravan - a global community dataset for large-sample hydrology,'' \emph{Scientific Data}, vol.~10, no.~1, p.~61, Jan. 2023. [Online]. Available: \url{https://doi.org/10.1038/s41597-023-01975-w}
\BIBentrySTDinterwordspacing

\bibitem{hysets_2020}
\BIBentryALTinterwordspacing
R.~Arsenault, F.~Brissette, J.-L. Martel, M.~Troin, G.~Lévesque, J.~Davidson-Chaput, M.~C. Gonzalez, A.~Ameli, and A.~Poulin, ``A comprehensive, multisource database for hydrometeorological modeling of 14,425 {North} {American} watersheds,'' \emph{Scientific Data}, vol.~7, no.~1, p. 243, Jul. 2020. [Online]. Available: \url{https://doi.org/10.1038/s41597-020-00583-2}
\BIBentrySTDinterwordspacing

\bibitem{lamah-2021}
\BIBentryALTinterwordspacing
C.~Klingler, K.~Schulz, and M.~Herrnegger, ``Lamah-ce: Large-sample data for hydrology and environmental sciences for central europe,'' \emph{Earth System Science Data}, vol.~13, no.~9, pp. 4529--4565, 2021. [Online]. Available: \url{https://essd.copernicus.org/articles/13/4529/2021/}
\BIBentrySTDinterwordspacing

\bibitem{era5-2021}
\BIBentryALTinterwordspacing
J.~Mu\~noz Sabater, E.~Dutra, A.~Agust\'{\i}-Panareda, C.~Albergel, G.~Arduini, G.~Balsamo, S.~Boussetta, M.~Choulga, S.~Harrigan, H.~Hersbach, B.~Martens, D.~G. Miralles, M.~Piles, N.~J. Rodr\'{\i}guez-Fern\'andez, E.~Zsoter, C.~Buontempo, and J.-N. Th\'epaut, ``Era5-land: a state-of-the-art global reanalysis dataset for land applications,'' \emph{Earth System Science Data}, vol.~13, no.~9, pp. 4349--4383, 2021. [Online]. Available: \url{https://essd.copernicus.org/articles/13/4349/2021/}
\BIBentrySTDinterwordspacing

\bibitem{gsim-p1-2018}
\BIBentryALTinterwordspacing
H.~X. Do, L.~Gudmundsson, M.~Leonard, and S.~Westra, ``The global streamflow indices and metadata archive (gsim) -- part 1: The production of a daily streamflow archive and metadata,'' \emph{Earth System Science Data}, vol.~10, no.~2, pp. 765--785, 2018. [Online]. Available: \url{https://essd.copernicus.org/articles/10/765/2018/}
\BIBentrySTDinterwordspacing

\bibitem{gsim-p2-2018}
\BIBentryALTinterwordspacing
L.~Gudmundsson, H.~X. Do, M.~Leonard, and S.~Westra, ``The global streamflow indices and metadata archive (gsim) -- part 2: Quality control, time-series indices and homogeneity assessment,'' \emph{Earth System Science Data}, vol.~10, no.~2, pp. 787--804, 2018. [Online]. Available: \url{https://essd.copernicus.org/articles/10/787/2018/}
\BIBentrySTDinterwordspacing

\bibitem{hydroatlas-2019}
\BIBentryALTinterwordspacing
S.~Linke, B.~Lehner, C.~Ouellet~Dallaire, J.~Ariwi, G.~Grill, M.~Anand, P.~Beames, V.~Burchard-Levine, S.~Maxwell, H.~Moidu, F.~Tan, and M.~Thieme, ``Global hydro-environmental sub-basin and river reach characteristics at high spatial resolution,'' \emph{Scientific Data}, vol.~6, no.~1, p. 283, Dec. 2019. [Online]. Available: \url{https://doi.org/10.1038/s41597-019-0300-6}
\BIBentrySTDinterwordspacing

\bibitem{nldas-2012}
Y.~Xia, K.~Mitchell, M.~Ek, J.~Sheffield, B.~Cosgrove, E.~Wood, L.~Luo, C.~Alonge, H.~Wei, J.~Meng, B.~Livneh, D.~Lettenmaier, V.~Koren, Q.~Duan, K.~Mo, Y.~Fan, and D.~Mocko, ``\BIBforeignlanguage{English (US)}{Continental-scale water and energy flux analysis and validation for the north american land data assimilation system project phase 2 (nldas-2): 1. intercomparison and application of model products},'' \emph{\BIBforeignlanguage{English (US)}{Journal of Geophysical Research Atmospheres}}, vol. 117, no.~3, 2012, copyright: Copyright 2018 Elsevier B.V., All rights reserved.

\bibitem{daymet-2014}
\BIBentryALTinterwordspacing
P.~THORNTON, M.~THORNTON, B.~MAYER, N.~WILHELMI, Y.~WEI, R.~DEVARAKONDA, and R.~COOK, ``\BIBforeignlanguage{en}{Daymet: Daily surface weather data on a 1-km grid for north america, version 2},'' 2014. [Online]. Available: \url{http://daac.ornl.gov/cgi-bin/dsviewer.pl?ds\_id=1219}
\BIBentrySTDinterwordspacing

\bibitem{lstm-1997}
S.~Hochreiter and J.~Schmidhuber, ``Long short-term memory,'' \emph{Neural Computation}, vol.~9, no.~8, pp. 1735--1780, 1997.

\bibitem{pidwirny_hydrologic_cycle}
\BIBentryALTinterwordspacing
M.~Pidwirny, ``The hydrologic cycle,'' 2006. [Online]. Available: \url{http://www.physicalgeography.net/fundamentals/8b.html}
\BIBentrySTDinterwordspacing

\bibitem{kingma2017adammethodstochasticoptimization}
\BIBentryALTinterwordspacing
D.~P. Kingma and J.~Ba, ``Adam: A method for stochastic optimization,'' 2017. [Online]. Available: \url{https://arxiv.org/abs/1412.6980}
\BIBentrySTDinterwordspacing

\bibitem{nnse-2012}
J.~{Nossent} and W.~{Bauwens}, ``{Application of a normalized Nash-Sutcliffe efficiency to improve the accuracy of the Sobol' sensitivity analysis of a hydrological model},'' in \emph{EGU General Assembly Conference Abstracts}, ser. EGU General Assembly Conference Abstracts, Apr. 2012, p. 237.

\bibitem{doi:10.1080/02626669509491401}
\BIBentryALTinterwordspacing
H.~RAMAN and N.~SUNILKUMAR, ``Multivariate modelling of water resources time series using artificial neural networks,'' \emph{Hydrological Sciences Journal}, vol.~40, no.~2, pp. 145--163, 1995. [Online]. Available: \url{https://doi.org/10.1080/02626669509491401}
\BIBentrySTDinterwordspacing

\bibitem{WOLD198737}
\BIBentryALTinterwordspacing
S.~Wold, K.~Esbensen, and P.~Geladi, ``Principal component analysis,'' \emph{Chemometrics and Intelligent Laboratory Systems}, vol.~2, no.~1, pp. 37--52, 1987, proceedings of the Multivariate Statistical Workshop for Geologists and Geochemists. [Online]. Available: \url{https://www.sciencedirect.com/science/article/pii/0169743987800849}
\BIBentrySTDinterwordspacing

\bibitem{neuralforecast2022}
\BIBentryALTinterwordspacing
K.~G. Olivares, C.~Challú, F.~Garza, M.~M. Canseco, and A.~Dubrawski, ``Neuralforecast: User friendly state-of-the-art neuralforecasting models.'' {PyCon} Salt Lake City, Utah, US 2022, 2022. [Online]. Available: \url{https://github.com/Nixtla/neuralforecast}
\BIBentrySTDinterwordspacing

\bibitem{mirchandani2023largelanguagemodelsgeneral}
\BIBentryALTinterwordspacing
S.~Mirchandani, F.~Xia, P.~Florence, B.~Ichter, D.~Driess, M.~G. Arenas, K.~Rao, D.~Sadigh, and A.~Zeng, ``Large language models as general pattern machines,'' 2023. [Online]. Available: \url{https://arxiv.org/abs/2307.04721}
\BIBentrySTDinterwordspacing

\bibitem{vaswani2023attentionneed}
\BIBentryALTinterwordspacing
A.~Vaswani, N.~Shazeer, N.~Parmar, J.~Uszkoreit, L.~Jones, A.~N. Gomez, L.~Kaiser, and I.~Polosukhin, ``Attention is all you need,'' 2023. [Online]. Available: \url{https://arxiv.org/abs/1706.03762}
\BIBentrySTDinterwordspacing

\bibitem{tolstikhin2021mlpmixerallmlparchitecturevision}
\BIBentryALTinterwordspacing
I.~Tolstikhin, N.~Houlsby, A.~Kolesnikov, L.~Beyer, X.~Zhai, T.~Unterthiner, J.~Yung, A.~Steiner, D.~Keysers, J.~Uszkoreit, M.~Lucic, and A.~Dosovitskiy, ``Mlp-mixer: An all-mlp architecture for vision,'' 2021. [Online]. Available: \url{https://arxiv.org/abs/2105.01601}
\BIBentrySTDinterwordspacing

\bibitem{chen2023tsmixerallmlparchitecturetime}
\BIBentryALTinterwordspacing
S.-A. Chen, C.-L. Li, N.~Yoder, S.~O. Arik, and T.~Pfister, ``Tsmixer: An all-mlp architecture for time series forecasting,'' 2023. [Online]. Available: \url{https://arxiv.org/abs/2303.06053}
\BIBentrySTDinterwordspacing

\bibitem{das2023long}
A.~Das, W.~Kong, A.~Leach, S.~Mathur, R.~Sen, and R.~Yu, ``Long-term forecasting with tide: Time-series dense encoder,'' \emph{arXiv preprint arXiv:2304.08424}, 2023.

\bibitem{ansari2024chronoslearninglanguagetime}
\BIBentryALTinterwordspacing
A.~F. Ansari, L.~Stella, C.~Turkmen, X.~Zhang, P.~Mercado, H.~Shen, O.~Shchur, S.~S. Rangapuram, S.~P. Arango, S.~Kapoor, J.~Zschiegner, D.~C. Maddix, H.~Wang, M.~W. Mahoney, K.~Torkkola, A.~G. Wilson, M.~Bohlke-Schneider, and Y.~Wang, ``Chronos: Learning the language of time series,'' 2024. [Online]. Available: \url{https://arxiv.org/abs/2403.07815}
\BIBentrySTDinterwordspacing

\bibitem{ye2024surveytimeseriesfoundation}
\BIBentryALTinterwordspacing
J.~Ye, W.~Zhang, K.~Yi, Y.~Yu, Z.~Li, J.~Li, and F.~Tsung, ``A survey of time series foundation models: Generalizing time series representation with large language model,'' 2024. [Online]. Available: \url{https://arxiv.org/abs/2405.02358}
\BIBentrySTDinterwordspacing

\bibitem{gruver2024large}
N.~Gruver, M.~Finzi, S.~Qiu, and A.~G. Wilson, ``Large language models are zero-shot time series forecasters,'' \emph{Advances in Neural Information Processing Systems}, vol.~36, 2024.

\bibitem{lim2021temporal}
B.~Lim, S.~{\"O}. Ar{\i}k, N.~Loeff, and T.~Pfister, ``Temporal fusion transformers for interpretable multi-horizon time series forecasting,'' \emph{International Journal of Forecasting}, vol.~37, no.~4, pp. 1748--1764, 2021.

\bibitem{bai2018empirical}
S.~Bai, J.~Z. Kolter, and V.~Koltun, ``An empirical evaluation of generic convolutional and recurrent networks for sequence modeling,'' \emph{arXiv preprint arXiv:1803.01271}, 2018.

\bibitem{chang2017dilated}
S.~Chang, Y.~Zhang, W.~Han, M.~Yu, X.~Guo, W.~Tan, X.~Cui, M.~Witbrock, M.~A. Hasegawa-Johnson, and T.~S. Huang, ``Dilated recurrent neural networks,'' \emph{Advances in neural information processing systems}, vol.~30, 2017.

\bibitem{wu2022timesnet}
H.~Wu, T.~Hu, Y.~Liu, H.~Zhou, J.~Wang, and M.~Long, ``Timesnet: Temporal 2d-variation modeling for general time series analysis,'' \emph{arXiv preprint arXiv:2210.02186}, 2022.

\bibitem{liu2023itransformer}
Y.~Liu, T.~Hu, H.~Zhang, H.~Wu, S.~Wang, L.~Ma, and M.~Long, ``itransformer: Inverted transformers are effective for time series forecasting,'' \emph{arXiv preprint arXiv:2310.06625}, 2023.

\bibitem{nie2022time}
Y.~Nie, N.~H. Nguyen, P.~Sinthong, and J.~Kalagnanam, ``A time series is worth 64 words: Long-term forecasting with transformers,'' \emph{arXiv preprint arXiv:2211.14730}, 2022.

\bibitem{lai2018modeling}
G.~Lai, W.-C. Chang, Y.~Yang, and H.~Liu, ``Modeling long-and short-term temporal patterns with deep neural networks,'' in \emph{The 41st international ACM SIGIR conference on research \& development in information retrieval}, 2018, pp. 95--104.

\bibitem{makridakis2020m4}
S.~Makridakis, E.~Spiliotis, and V.~Assimakopoulos, ``The m4 competition: 100,000 time series and 61 forecasting methods,'' \emph{International Journal of Forecasting}, vol.~36, no.~1, pp. 54--74, 2020.

\bibitem{wu2021autoformer}
H.~Wu, J.~Xu, J.~Wang, and M.~Long, ``Autoformer: Decomposition transformers with auto-correlation for long-term series forecasting,'' \emph{Advances in neural information processing systems}, vol.~34, pp. 22\,419--22\,430, 2021.

\bibitem{JMLR:v21:20-074}
\BIBentryALTinterwordspacing
C.~Raffel, N.~Shazeer, A.~Roberts, K.~Lee, S.~Narang, M.~Matena, Y.~Zhou, W.~Li, and P.~J. Liu, ``Exploring the limits of transfer learning with a unified text-to-text transformer,'' \emph{Journal of Machine Learning Research}, vol.~21, no. 140, pp. 1--67, 2020. [Online]. Available: \url{http://jmlr.org/papers/v21/20-074.html}
\BIBentrySTDinterwordspacing

\bibitem{radford2019language}
A.~Radford, J.~Wu, R.~Child, D.~Luan, D.~Amodei, I.~Sutskever \emph{et~al.}, ``Language models are unsupervised multitask learners,'' 2019.

\bibitem{jin2024timellmtimeseriesforecasting}
\BIBentryALTinterwordspacing
M.~Jin, S.~Wang, L.~Ma, Z.~Chu, J.~Y. Zhang, X.~Shi, P.-Y. Chen, Y.~Liang, Y.-F. Li, S.~Pan, and Q.~Wen, ``Time-llm: Time series forecasting by reprogramming large language models,'' 2024. [Online]. Available: \url{https://arxiv.org/abs/2310.01728}
\BIBentrySTDinterwordspacing

\bibitem{Huang2019mlc}
\BIBentryALTinterwordspacing
X.~Huang, G.~C. Fox, S.~Serebryakov, A.~Mohan, P.~Morkisz, and D.~Dutta, ``Benchmarking deep learning for time series: Challenges and directions,'' in \emph{2019 IEEE International Conference on Big Data (Big Data)}.\hskip 1em plus 0.5em minus 0.4em\relax ieeexplore.ieee.org, Dec. 2019, pp. 5679--5682. [Online]. Available: \url{http://dx.doi.org/10.1109/BigData47090.2019.9005496}
\BIBentrySTDinterwordspacing

\bibitem{he_2024_13975174}
\BIBentryALTinterwordspacing
J.~He, ``Data for science time series: Deep learning in hydrology,'' Oct. 2024. [Online]. Available: \url{https://doi.org/10.5281/zenodo.13975174}
\BIBentrySTDinterwordspacing

\end{thebibliography}

\end{document}